\documentclass[10pt,twocolumn,letterpaper]{article}

\usepackage[pagenumbers]{cvpr} % To force page numbers, e.g. for an arXiv version
\usepackage{algorithm}
\usepackage{algorithmic}
\usepackage{listings}
\definecolor{cvprblue}{rgb}{0.21,0.49,0.74}
\newcommand{\smallsec}[1]{\vspace{3pt}\noindent\textbf{#1}}
\usepackage[pagebackref,breaklinks,colorlinks,allcolors=cvprblue]{hyperref}

\def\paperID{11001} % *** Enter the Paper ID here
\def\confName{CVPR}
\def\confYear{2025}

\title{Iterative Predictor-Critic Code Decoding for Real-World Image Dehazing}

\author{Jiayi Fu\textsuperscript{1}\quad Siyu Liu\textsuperscript{1}\quad Zikun Liu\textsuperscript{3}\quad Chun-Le Guo\textsuperscript{1,2}\quad \\
Hyunhee Park\textsuperscript{4}\quad 
Ruiqi Wu\textsuperscript{1}\quad Guoqing Wang\textsuperscript{5}\quad Chongyi Li\textsuperscript{1,2}\thanks{Corresponding author}\quad \\
\textsuperscript{1}VCIP, CS, Nankai University\quad
\textsuperscript{2}NKIARI, Shenzhen Futian \\ 
\textsuperscript{3}Samsung R\&D Institute China-Beijing\quad
\textsuperscript{4}CIG, Samsung Electronics\\
\textsuperscript{5}Donghai Laboratory, Zhoushan, Zhejiang\\
{\tt\small \{fujiayi,liusiyu29\}@mail.nankai.edu.cn, zikun.liu@samsung.com,}\\
{\tt\small  guochunle@nankai.edu.cn, inextg.park@samsung.com,}\\
{\tt\small wuruiqi@mail.nankai.edu.cn, gqwang0420@hotmail.com, lichongyi@nankai.edu.cn}\\
{\tt\small \href{https://github.com/Jiayi-Fu/IPC-Dehaze}{[Code]}\quad\href{https://jiayi-fu.github.io/IPC-Dehaze_Homepage/}{[Website]}
}
}

\begin{document}
\maketitle
\begin{abstract}
We propose a novel \textbf{I}terative \textbf{P}redictor-\textbf{C}ritic Code Decoding framework for real-world image dehazing, abbreviated as \textbf{IPC-Dehaze}, which leverages the high-quality codebook prior encapsulated in a pre-trained VQGAN. Apart from previous codebook-based methods that rely on one-shot decoding, our method utilizes high-quality codes obtained in the previous iteration to guide the prediction of the Code-Predictor in the subsequent iteration, improving code prediction accuracy and ensuring stable dehazing performance. Our idea stems from the observations that 1) the degradation of hazy images varies with haze density and scene depth, and 2) clear regions play crucial cues in restoring dense haze regions. However, it is non-trivial to progressively refine the obtained codes in subsequent iterations, owing to the difficulty in determining which codes should be retained or replaced at each iteration. Another key insight of our study is to propose Code-Critic to capture interrelations among codes. The Code-Critic is used to evaluate code correlations and then resample a set of codes with the highest mask scores, i.e., a higher score indicates that the code is more likely to be rejected, which helps retain more accurate codes and predict difficult ones. Extensive experiments demonstrate the superiority of our method over state-of-the-art methods in real-world dehazing.
\end{abstract}
\vspace{-0.5em}
\section{Introduction}
\label{sec:intro}
In real-world scenarios, capturing hazy images is a common occurrence, wherein the visual output exhibits diminished clarity and sharpness. This phenomenon is primarily attributable to the presence of atmospheric particles that irregularly disperse and scatter light. The presence of haze can significantly impact the visual quality or accuracy of subsequent perception algorithms. Consequently, there has been a growing interest in image dehazing, aiming to restore clear images from their hazy counterparts.
\begin{figure}[t]
    \centering
    \includegraphics[width=0.235\textwidth]{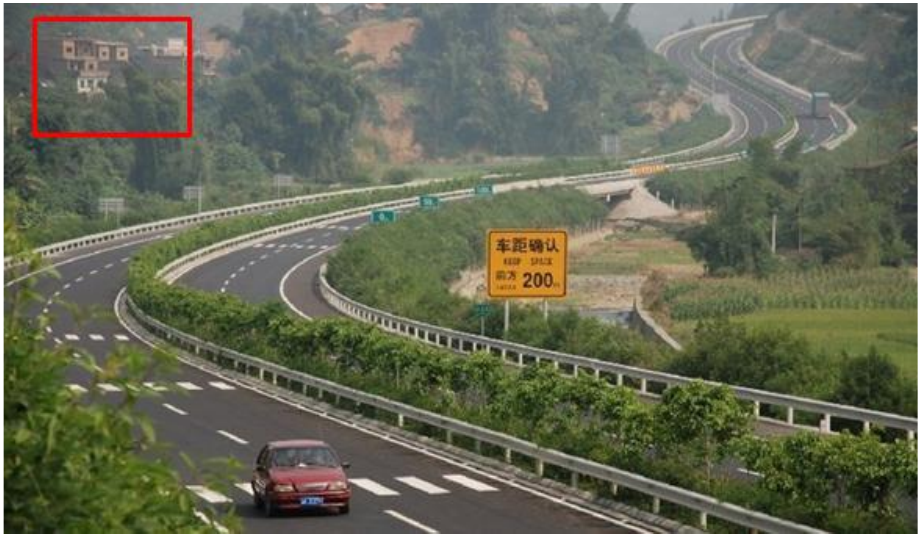}
    \includegraphics[width=0.235\textwidth]{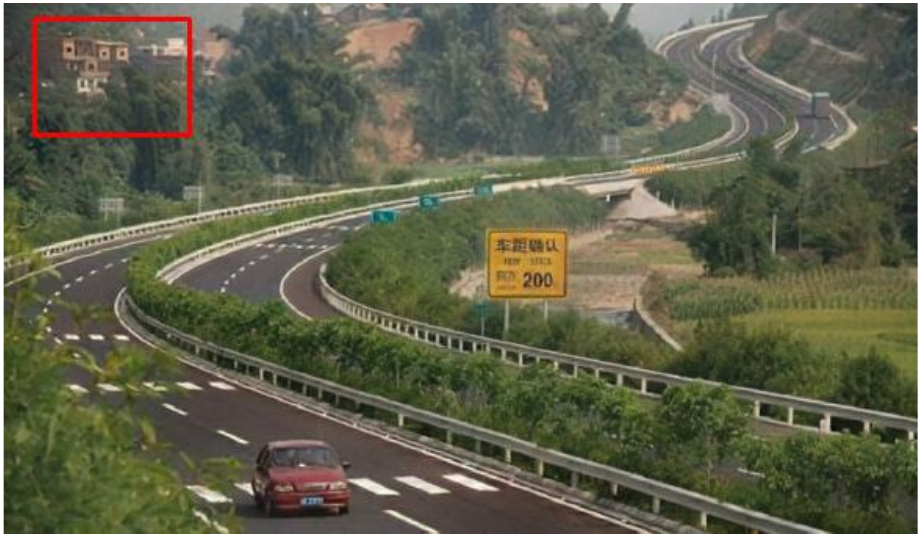}
     \\
     \vspace{-0.2em}
    \makebox[0.235\textwidth]{\small Hazy image}
    \makebox[0.235\textwidth]{\small RIDCP~\cite{Wu_2023_CVPR}}

    \includegraphics[width=0.235\textwidth]{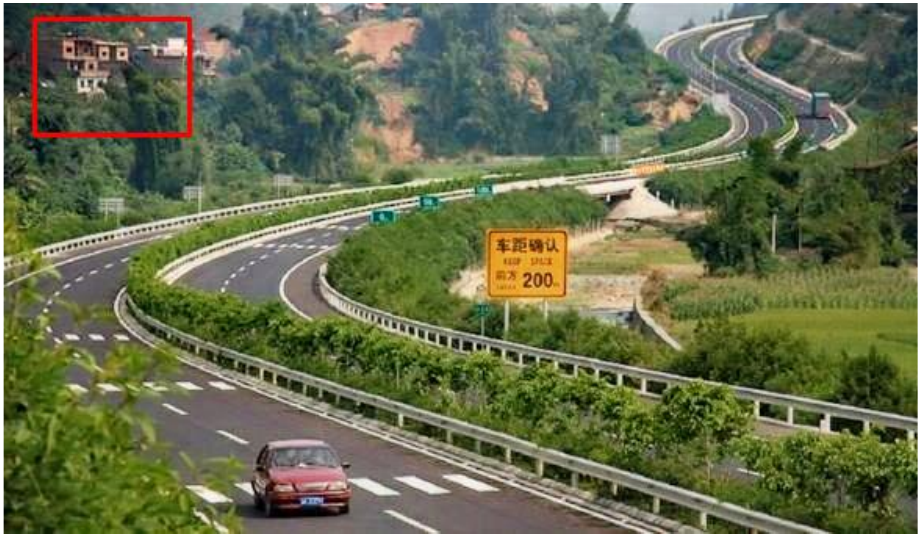}
    \includegraphics[width=0.235\textwidth]{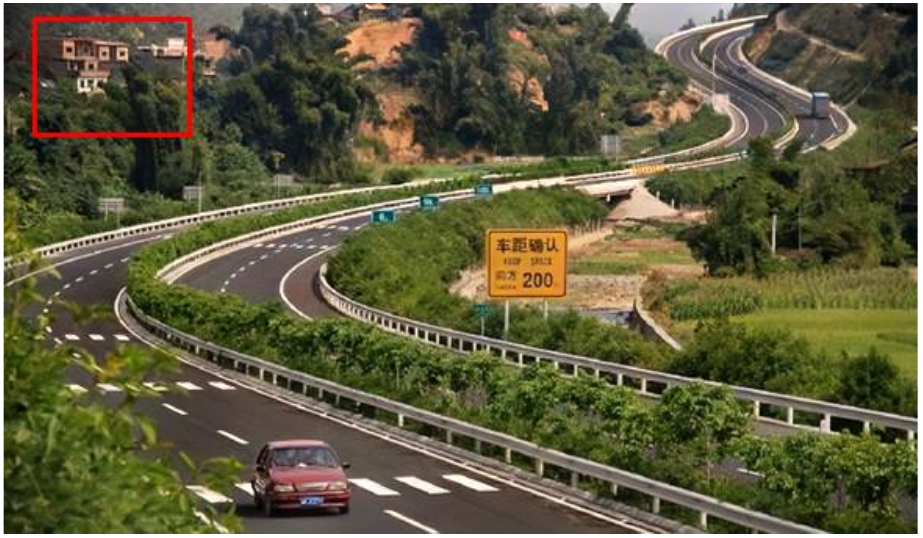}
     \\
     \vspace{-0.2em}
    \makebox[0.235\textwidth]{\small KA-Net~\cite{KA}}
    \makebox[0.235\textwidth]{\small IPC-Dehaze (Ours)}
    \\
     \vspace{0.2em}
     \includegraphics[width=0.48\textwidth]{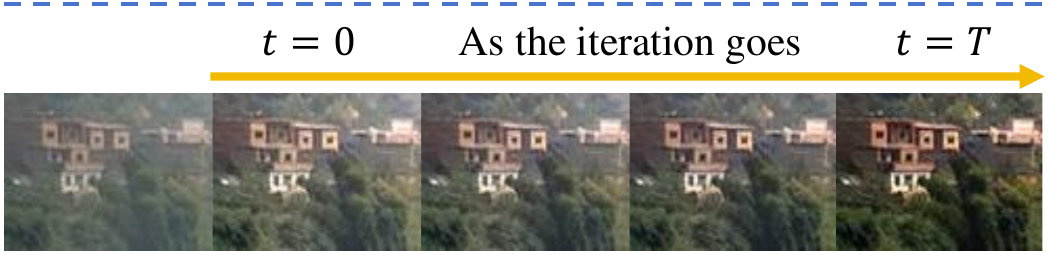}
     \vspace{-0.2em}
     \makebox[=0.48\textwidth]{Iterative results of our IPC-Dehaze}
    \caption{A comparison between the state-of-the-art real-world image dehazing methods and our IPC-Dehaze. In comparison, our result is sharper and clearer, with less color distortion and overexposure. The bottom images present the results of each iteration in our method, showing the continuous improvements with our core Predictor-Critic mechanism.}
    \vspace{-0.5em}
    \label{fig:teaser}
\end{figure}

Achieving a clean image from a hazy one is inherently challenging. Traditional methods address this ill-posed problem by leveraging various priors~\cite{Fattal_2014,he2010single, Zhu_Mai_Shao_2014, Berman_Treibitz_Avidan_2016}, but they often struggle in real-world scenes due to ideal assumptions or limited generalization capabilities.
The advent of deep learning has led to data-driven methods exhibiting remarkable performance in image dehazing. These methods either estimate the physical model-related transmission map~\cite{Cai_Xu_Jia_Qing_Tao_2016, Ren_Liu_Zhang_Pan_Cao_Yang_2016} or directly restore the clean image~\cite{Dong_Pan_Xiang_Hu_Zhang_Wang_Yang_2020, Qin_Wang_Bai_Xie_Jia_2020}. However, they are commonly trained on images synthesized by ideal physical models, posing challenges when applied to complex real-world scenes.

To address this issue, some approaches focus on improving data synthesis~\cite{Shao_Li_Ren_Gao_Sang_2020, Yang_Wang_Liu_Zhang_Guo_Tao} or employ alternative training strategies like unsupervised learning~\cite{Zhao_Zhang_Shen_Zhou_2021, Golts_Freedman_Elad_2020}. Recent work, RIDCP~\cite{Wu_2023_CVPR}, highlights the importance of considering multiple degradation factors for synthesizing hazy images and demonstrates the effectiveness of high-quality codebook priors. In line with RIDCP, some works ~\cite{Chen_Wang_Yang_Liu, Li_Dong_Ren_Pan_Gao_Sang_Yang_2020} have been proposed for real-world dehazing.
Despite the progress made by current methods, they still face some limitations. Specifically, they may overlook the spatial variations in degradation with haze density and scene depth, leading to over-recovery in thin haze areas or under-recovery in dense haze areas. Additionally,  accurately recovering a challenging scene with a one-shot (i.e., only a single attempt or instance)  mapping proves difficult, resulting in suboptimal performance.

Inspired by the physical phenomenon that areas with thin haze contain more information while areas with dense haze contain less, we found that dehazing efforts could benefit from focusing on relatively clear scenes first, before addressing more challenging areas. To achieve that, we follow previous methods~\cite{Wu_2023_CVPR} to utilize the high-quality codebook encapsulated in a pre-train VQGAN~\cite{vqgan} as prior. Unlike previous codebook-based methods that perform a one-shot code prediction, we present a novel iterative decoding prediction framework, which performs dehazing in an easy-to-hard manner, improving the overall accuracy and stability of the dehazing process, as the comparison shown in \cref{fig:teaser}.

In our method, an encoder first maps a hazy image into tokens. Then, according to the current tokens, a Predictor-Critic mechanism is used to alternately predict high-quality codes and evaluate which should be retained. As the iterations progress, the number of retained codes progressively grows until all codes are finalized. In particular, in the $t$-th iteration, we get the high-quality codes predicted in the previous iteration and a mask map that determines which codes
are retained in this iteration. Next, we feed the mask map processed tokens into the Code-Predictor to predict all the high-quality codes in this iteration, and we also need to generate a mask map to benefit the next iteration. To achieve the iterative dehazing shown above, we fuse the tokens from clean and hazy images with a random mask. The processed
tokens are fed to Code-Predictor to match the high-quality codebook, simulating the prediction of the $t$-th iteration during inference. Following the inference process, obtaining an effective and instructive mask map is important, so we propose Code-Critic to better reject the least likely codes and retain the good ones.

The \textbf{contributions} of this study are summarized as follows:
\begin{itemize}
\item We propose a novel iterative decoding dehazing framework. In comparison to one-shot methods, our method uses high-quality codes from previous iterations as cues to guide Code-Predictor in predicting the subsequent codes, implementing better iterative dehazing.

\item We introduce Code-Critic to evaluate the interdependence in the output of the Code-Predictor, selecting which codes should be retained or rejected across iterative decoding steps to guide subsequent predictions. This module improves the coherence among selected codes and prevents error accumulation.
\item With these novel designs, our approach surpasses state-of-the-art methods in real-world dehazing both qualitatively and quantitatively.
\end{itemize}

\section{Related Work}

Early single image dehazing methods ~\cite{Fattal_2014,he2010single,Zhu_Mai_Shao_2014,Berman_Treibitz_Avidan_2016} recover images by estimating the parameters of imaging model~\cite{Hide_1977}. However, hand-crafted priors suffer from poor generalization ability and cannot adapt to the various scenes. With the development of deep learning, data-driven methods have achieved remarkable results. One kind of approach~\cite{Cai_Xu_Jia_Qing_Tao_2016, Ren_Liu_Zhang_Pan_Cao_Yang_2016, Li_Peng_Wang_Xu_Feng_2017} is inspired by the scattering model, but the performance is unsatisfactory when the scene does not conform to the ideal physical model. Another line ~\cite{Qin_Wang_Bai_Xie_Jia_2020, Guo_Yan_Anwar_Cong_Ren_Li} is to directly obtain a clean image using a network. However, due to the domain gap between synthetic data and real data, the model trained on synthetic images often suffers from sub-optimal performance when applied to real-world hazy images.  

\begin{figure*}[thbp]
  \centering
  \includegraphics[width=\textwidth]{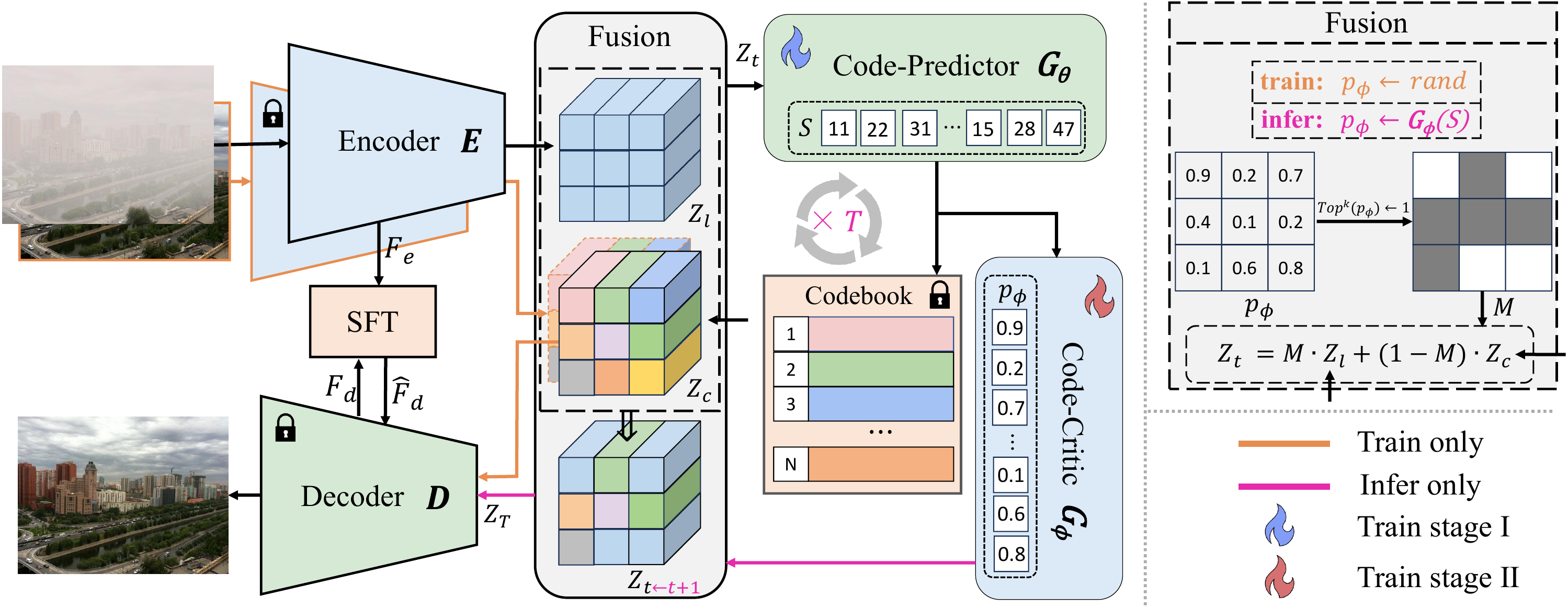}
  \caption{Overview of our IPC-Dehaze. \textit{In the training phase}, we use fused tokens $Z_{t}= Z_l \odot M + Z_c \odot (1-M)$ from the hazy and clean images, and predict the sequence codes $S$ by Code-Predictor. We also train Code-Critic to evaluate each code in set $S$ for potential rejection and resampling. \textit{In the inference phase}, $Z_{t=0}$ is initially encoded as low-quality tokens $Z_l$. During the $t$-th iterative decoding step, the Code-Predictor takes $Z_t$ as input, predicting the sequence codes $S$ and the corresponding high-quality tokens $Z_c$. To retain the reliable codes and resample the others, the Code-Critic evaluates $S$ and produces a mask map $M$ by $p_\phi$. This mask map $M$ is then used to generate $Z_{t+1}$ through a \emph{Fusion} process. Following $T$ iterations, $Z_T$ is output to reconstruct the clean image by a decoder. The SFT refers to the Spital Feature Transform, which adjusts the feature within the encoder and decoder. 
}
\vspace{-0.4em}
  \label{fig:pipeline}
\end{figure*}

There has been a remarkable amount of research focusing on real-world image dehazing. 
Real-world image dehazing methods can be roughly divided into three categories:
1) Domain Adaptation-based Method. Some methods employ domain adaptation to reduce the gap between the synthetic domain and the real domain ~\cite{Shao_Li_Ren_Gao_Sang_2020, Shyam_Yoon_Kim_2021}. These supervised methods typically achieve excellent performance on synthetic datasets, but they often struggle to generalize well to real-world hazy images. 
%overfit the provided training data and 
2) GAN-based Method. In contrast, GAN-based unsupervised dehazing methods learn the dehazing mapping from unpaired clean and hazy images. To produce clearer and more realistic images,
 Liu \etal~\cite{Liu_Hou_Duan_Qiu_2020} proposed a dehazing network based on a physical model and an enhancer network that supervises the mapping from the hazy domain to the clean domain. Yang \etal~\cite{Yang_Wang_Liu_Zhang_Guo_Tao} introduced an unpaired image dehazing framework that utilizes self-enhancement to re-render blurred images with various haze densities. However, GAN-based methods often tend to produce unrealistic haze, which further compromises the overall dehazing performance. 
3) Deep Prior-based Method. Some methods~\cite{Li_Dong_Ren_Pan_Gao_Sang_Yang_2020, Chen_Wang_Yang_Liu} that introduce handcrafted priors in the network framework or loss functions still fail to overcome their inherent limitations. Some efforts have been made to utilize prior knowledge extracted from high-quality images to aid in image restoration. Some face restoration~\cite{Zhou_Chan_Li_Loy_2022,gu2022vqfr} and super-resolution~\cite{chen2022real,chen2024iter} methods based on VQGAN~\cite{vqgan} have demonstrated the superiority of high-quality priors. RIDCP~\cite{Wu_2023_CVPR} leverage a latent high-quality prior in image dehazing and achieve significant performance. However, the one-shot-based algorithm struggles with extremely dense haze and performs poorly in the inhomogeneous haze, failing to take full advantage of the information in the thin haze. To address these issues, our new framework introduces generative capability and high-quality priors and enhances the model’s generalization through an iterative approach.

\section{Methodology}

In this work, we propose a novel image dehazing framework via iterative decoding, which is shown in \cref{fig:pipeline}. Firstly, to introduce robust high-quality priors, we pre-train a VQGAN~\cite{vqgan} on high-quality datasets. In this stage, we can obtain a latent high-quality discrete codebook and the corresponding encoder and decoder. To match the correct code, we use the proposed Code-Predictor replacing the nearest neighbor matching (Stage \uppercase\expandafter{\romannumeral1}). To capture the global relationship of the code sequence, we introduce Code-Critic for selecting whether a code generated by the Code-Predictor is accepted or not during iterative decoding (Stage \uppercase\expandafter{\romannumeral2}). This ensures that the
acceptance or rejection of a code should be not decided independently. \underline{We present the training process in \cref{algorithm:1}.}
During the inference phase, the process begins with tokens encoded from a hazy image. Over $T$ iterations, the current token is fed into the Code-Predictor and the Code-Critic selects which high-quality codes will be retained for the next iteration. This process continues until all codes have been replaced with high-quality ones. \underline{We present the inference
process in \cref{algorithm:2}.}

\subsection{Preliminary}
\smallsec{Learning A Codebook via VQGAN.} 
%GAN-based methods have achieved amazing success in image-generation tasks. 
In order to alleviate the complexity of directly generating images in pixel space, VQVAE~\cite{vqvae} proposes a vector quantized (VQ) autoencoder to learn discrete codebooks in the latent space. VQGAN~\cite{vqgan} further enhances the perceptual quality of the reconstruction results by introducing adversarial loss and perceptual loss. Following VQGAN, the codebook learning consists of three parts: 

\begin{enumerate}[i)]
\item The encoder $E_H$ encodes the high-quality image patch $I_h\in \mathbb{R}^{H\times W \times 3}$ to latent features $Z_h = E_H(I_h) \in \mathbb{R}^{m\times n \times d}$, where the $d$ denotes the dimension of codebook embedding vector.
\item Each item in $Z_h$ is replaced with the closest code in the codebook $C \in \mathbb{R}^{K\times d}$ with $K$ codebook size to obtain the quantized feature $Z_c \in \mathbb{R}^{m\times n \times d}$ and sequence $S_h$, which is formulated by \cref{eq:vq}:
\begin{equation}
\setlength\abovedisplayskip{3pt}%shrink space
\setlength\belowdisplayskip{3pt}
\begin{split}
Z_c^{(i,j)}=\mathbf{q}(Z_l)=\mathop{\arg\min}\limits_{c_k \in C} \Vert Z_h^{(i,j)} - c_k \Vert_2,
\\
S_h^{(i,j)} =k \    \text{such that} \ Z_c^{(i,j)}=c_k. 
\end{split}
\label{eq:vq}
\end{equation}
\item Reconstructing the image $I_h$ by the decoder $D_H$:
$I_{rec}=D_H(Z_c)$.
\end{enumerate}

\textbf{Analysis.} To reduce the ill-posedness of image dehazing, we pre-train a VQGAN to learn a latent discrete high-quality
codebook, which improves the network’s robustness against
various degradations. Due to the domain gap between the
hazy images and clean images, the nearest neighbor matching (\cref{eq:vq}) is ineffective in accurately matching codes.
This issue will be further discussed in the ablation study (see \cref{fig:prediction}). Thus, we will introduce Code-Predictor to better match codes.
% \subsection{Preliminary:MaskGIT(An Iterative Generation Paradigm).} 

\begin{algorithm}[t]
  \caption{Training Process}
  \label{algorithm:1}
    \textbf{Input}: $Z_l$, $Z_c$, $S_h$, Code-Predictor $G_\theta$, Code-Critic $G_\phi$, mask schedule functions $\gamma(r)$, learning rate $\eta$, number of tokens in latent $N$.
    
  \begin{algorithmic}[1]
    \REPEAT
    
    \STATE $r \sim \mathcal{U}_{(0,1)}$ 
    \STATE $M_t \leftarrow \text{randomly sample}(\lceil \gamma(r) \cdot N \rceil)$
    
    \STATE $Z_t \leftarrow Z_l \odot M_t + Z_c \odot (1 - M_t)$ \COMMENT{Confuse $Z_l$ and $Z_c$ based on $M_t$}
    \STATE $p_\theta \leftarrow G_\theta(Z_t)$ 
    \STATE $S \leftarrow \text{argmax}(p_\theta)$ 
    \STATE $\theta \leftarrow \theta - \eta \nabla_\theta \mathcal{L}_\theta$ 

      \IF{training Code-Critic}
        \STATE $S \leftarrow \text{sample from } p_\theta$
        \STATE $M \leftarrow (S \neq S
        _h)$ \COMMENT{Build ground truth}
        \STATE $\phi \leftarrow \phi - \eta \nabla_\phi \mathcal{L}_\phi$ 
      \ENDIF
    \UNTIL{convergence}
  \end{algorithmic}
\end{algorithm}

\begin{algorithm}[t]
  \caption{Inference Process}
  \label{algorithm:2}
  \textbf{Input}: Encoder $E_L$, decoder $D_H$, Code-Predictor $G_\theta$, Code-Critic $G_\phi$, mask schedule functions $\gamma(r)$, codebook $C$, number of tokens in latent $N$, number of iterations $T$, low-quality images $I_l$\\
  \textbf{Output}: clear images $I_{rec}$
  
  \begin{algorithmic}[1]
    \STATE $Z_l \leftarrow E_L(I_l)$
    \STATE $M_1 \leftarrow 1$ \COMMENT{$M_1$ initialized to 1}
    % \STATE $S \leftarrow \text{sample from } G_\theta(Z_l)$ \COMMENT{$G_{\theta}$ outputs the set of probability distributions}
    % \STATE $Z_c \leftarrow C(S)$ \COMMENT{Using $S$ to select tokens from the Codebook}
    \FOR{$t \leftarrow 1$ \TO $T$}
      \STATE $Z_t \leftarrow Z_l \odot M_t + Z_c \odot (1 - M_t)$
      \STATE $p_\theta \leftarrow G_\theta(Z_t)$
      \STATE $S \leftarrow \text{sample from } p_\theta $
      \STATE $Z_c \leftarrow C(S)$ \COMMENT{Using $S$ to select tokens from the codebook $C$}
       \STATE $p_\phi \leftarrow G_\phi(S)$
      \STATE $M_{t+1} \leftarrow \text{sample} \lceil \gamma(\frac{t}{T}) \cdot N \rceil \text{ from } p_\phi$ 
    \ENDFOR
    \STATE $I_{rec} \leftarrow D_H(Z_c)$
  \end{algorithmic}
\vspace{-0.2em}
\end{algorithm}

\smallsec{MaskGIT: An Iterative Generation Paradigm.} MaskGIT~\cite{maskgit} is an image generation approach that incorporates masked token modeling and parallel decoding, improving both performance and efficiency. Let $Y=[y_{i}]_{i=1}^{N}$ denote the latent tokens derived from the input image via an encoder, $M$ represents the corresponding binary mask of $Y$, and $Y_{\bar{M}}$ represents $Y$ masked by $M$. During training, MaskGIT trains a predictor that can predict the masked parts through $Y_{\bar{M}}$, aiming to maximize the posterior probability $P(y_{i}|Y_{\bar{M}})$. During inference, MaskGIT predicts all tokens, keeping only the most confident ones, and re-sampling the rest in the next iteration.

\textbf{Analysis.} In contrast to image generation, image restoration emphasizes the fidelity of the original scene. Thus, it makes sense to use the low-quality features of the image as a condition. Directly applying MaskGIT to image restoration is infeasible as it is designed for LQ-to-HQ mapping rather than HQ-to-HQ mapping; meanwhile, the mechanism for independently sampling tokens ignores their interdependence~\cite{critic}. To address these issues, we employ Code-Critic at each iteration to identify the relationships between tokens. This approach effectively identifies which tokens to mask at each iteration.

\begin{table*}[htpb]
    \centering
    \caption{Quantitative comparison using NR-IQA on RTTS, Fattal, and URHI datasets. \textcolor{red}{Red} and \textcolor{blue}{blue} indicate the best, second-best performers respectively. $^{~\star}$ indicate that due to image size constraints, we are unable to test the results under CLIPIQA and TOPIQ metrics.}
    \label{tab:1}
  \resizebox{\textwidth}{!}{
      \Huge
    \begin{tabular}{c|c|cccc|cccccc}
    \toprule
    \multicolumn{1}{l|}{Datasets} & Metrics & Hazy images & MSBDN~\cite{Dong_Pan_Xiang_Hu_Zhang_Wang_Yang_2020} & Dehamer~\cite{Guo_Yan_Anwar_Cong_Ren_Li} & DEA-Net~\cite{DEA} & DAD~\cite{Shao_Li_Ren_Gao_Sang_2020}   & D4~\cite{Yang_Wang_Liu_Zhang_Guo_Tao}    & PSD~\cite{Chen_Wang_Yang_Liu}   & RIDCP~\cite{Wu_2023_CVPR} & KA-Net~\cite{KA} & Ours \\
    \midrule
          & MUSIQ$\uparrow$ & 53.76 & 53.73 & 53.57  & 54.09 & 49.33 & 53.55 & 50.30  & \textcolor{blue}{55.23} & 54.64 & \textcolor[rgb]{ 1,  0,  0}{59.60} \\
          & PI$\downarrow$   & 4.78  & 4.15  & 4.41  & 3.83  & 4.19  & 3.86  & 3.61  & \textcolor[rgb]{ 0,  0,  1}{3.56} & 3.63  & \textcolor[rgb]{ 1,  0,  0}{3.22} \\
    RTTS  & MANIQA$\uparrow$ & 0.311 & 0.311 & 0.310  & \textcolor[rgb]{ 0,  0,  1}{0.314} & 0.221 & 0.297 & 0.256  & 0.251 & 0.259 & \textcolor[rgb]{ 1,  0,  0}{0.327} \\
          & CLIPIQA$\uparrow$ & \textcolor[rgb]{ 0,  0,  1}{0.39} & 0.36  & 0.36  & 0.37  & 0.25  & 0.34  & 0.28  & 0.30  & 0.28  & \textcolor[rgb]{ 1,  0,  0}{0.44} \\
          & Q-Align$\uparrow$ & 3.04  & 3.04  & 3.10  & 3.11  & 2.84  & 2.97  & 2.63  & \textcolor[rgb]{ 0,  0,  1}{3.24} & 3.09  & \textcolor[rgb]{ 1,  0,  0}{3.49} \\
          & TOPIQ$\uparrow$ & 0.400  & 0.406  & 0.401  & 0.407  & 0.335  & 0.402  & 0.353  & \textcolor[rgb]{ 0,  0,  1}{0.412} & 0.394  & \textcolor[rgb]{ 1,  0,  0}{0.500} \\

    \midrule
          & MUSIQ$\uparrow$ & 63.61 & 63.67 & 64.40 & 63.33 & 58.17 & 63.92 & 60.96  & \textcolor[rgb]{ 0,  0,  1}{65.48} & 64.09 & \textcolor[rgb]{ 1,  0,  0}{66.22} \\
          & PI$\downarrow$   & 3.18  & 2.47  & 2.48  & 2.66  & 3.02  & 2.44  & 2.83  & \textcolor[rgb]{ 1,  0,  0}{2.37} & 2.81  & \textcolor[rgb]{ 0,  0,  1}{2.41} \\
    Fattal & MANIQA$\uparrow$ & 0.38  & 0.38  & 0.38  & \textcolor[rgb]{ 0,  0,  1}{0.40} & 0.26  & 0.39  & 0.33  & 0.31  & 0.35  & \textcolor[rgb]{ 1,  0,  0}{0.43} \\
          & CLIPIQA$\uparrow$ & 0.51  & 0.53  & 0.51  & 0.50  & 0.42  & \textcolor[rgb]{ 0,  0,  1}{0.55} & 0.48  & 0.42  & 0.50  & \textcolor[rgb]{ 1,  0,  0}{0.59} \\
          & Q-Align$\uparrow$ & 3.739  & 3.943  & 3.923  & 3.934  & 3.593  & 3.980  & 3.312  & 3.799  & \textcolor[rgb]{ 0,  0,  1}{3.982} & \textcolor[rgb]{ 1,  0,  0}{4.234} \\
          & TOPIQ$\uparrow$ & 0.56  & 0.56  & 0.56  & 0.58  & 0.41  & \textcolor[rgb]{ 0,  0,  1}{0.59} & 0.49  & 0.50  & 0.55  & \textcolor[rgb]{ 1,  0,  0}{0.63} \\
          
\midrule
          & MUSIQ$\uparrow$ & 57.8  & 57.35 & 57.08 & 57.64 & 57.12 & 52.23 & 53.96  & \textcolor[rgb]{ 0,  0,  1}{61.39} & 58.57 & \textcolor[rgb]{ 1,  0,  0}{62.5} \\
    URHI$^{\star}$  & PI$\downarrow$   & 4.07  & 3.57  & 3.88  & 3.83  & 3.64  & 3.71  & 3.48  & \textcolor[rgb]{ 1,  0,  0}{2.87} & 3.15  & \textcolor[rgb]{ 0,  0,  1}{3.08} \\
          & MANIQA$\uparrow$ & 0.355  & 0.348  & 0.350  & \textcolor[rgb]{ 0,  0,  1}{0.355} & 0.344  & 0.262  & 0.294  & 0.314  & 0.306  & \textcolor[rgb]{ 1,  0,  0}{0.364} \\
          & Q-Align$\uparrow$ & 3.21  & 3.19  & 3.10  & 3.28  & 2.92  & 3.16  & 2.68 & \textcolor[rgb]{ 0,  0,  1}{3.51}  & 3.23  & \textcolor[rgb]{ 1,  0,  0}{3.70} \\
    \bottomrule
    \end{tabular}%
    }
\end{table*}%

\subsection{Code-Predictor Training (Stage \uppercase\expandafter{\romannumeral1})}
\label{sec:stage2}

 In this stage, we will fix codebook $C$, decoder $D_H$, and pre-train encoder $E_H$ (also referred to as $E_L$). When we have haze images $I_l$, we can obtain corresponding features $Z_l$ with $E_L(I_l)$. We then randomly sample a binary mask matrix $M_t \in \mathbb{R}^{m \times n} $, where the mask ratio can be determined by \(\left\lceil \gamma(r) \cdot (m \times n) \right\rceil\), where \(\gamma(r)\) denotes the cosine function, and \(r\) is a random number sampled from a Uniform(0,1]. To obtain the input $Z_t$, we apply $M$ to $Z_l$ and $Z_c$, following the formulated by:
\begin{equation}
\setlength\abovedisplayskip{3pt}%shrink space
\setlength\belowdisplayskip{3pt}
    Z_t = Z_l \odot M_t + Z_c \odot (1-M_t).
\label{eq:mask}
\end{equation}

Then, $Z_t$ is fed into Code-Predictor  $G_\theta$ to forecast the probability $p_\theta \in \mathbb{R}^{(m\times n)\times K}$ over all codes in the codebook $C$. We adopt cross-entropy loss $\mathcal{L}_\theta$ to train $G_\theta$:
\begin{align}
\setlength\abovedisplayskip{3pt}%shrink space
\setlength\belowdisplayskip{3pt}
    \mathcal{L}_\theta = -\sum_{i=0}^N S_h^{(i)} \log p_\theta(S_h^{(i)}|Z_t),
\end{align}
where $i$ is the index, $N$ is the number of elements.

To align the features between hazy images and high-quality images, we introduce the SFT  module~\cite{sft, Zhou_Chan_Li_Loy_2022}:
\begin{equation}
\setlength\abovedisplayskip{3pt}%shrink space
\setlength\belowdisplayskip{3pt}
    \hat{F}_d=F_d+\alpha \odot F_d+\beta; \alpha, \beta =Conv(concat(F_e,F_d)).
\end{equation}

We jointly train low-quality image encoder $E_L$, Code-Predictor $G_\theta$, and SFT. Further training details will be provided in the supplementary material.

\subsection{Code-Critic Training (Stage  \uppercase\expandafter{\romannumeral2})}
\label{sec:stage3}

In training Stage \uppercase\expandafter{\romannumeral1}, we obtain high-quality restored images. However, during the inference stage, we only sample codes $S$ from the output distribution $p_\theta$ of the Code-Predictor, without considering the relationships between the codes. To overcome this challenge, in Stage \uppercase\expandafter{\romannumeral2} training, we keep other model components fixed and introduce the Code-Critic $G_\phi$ to evaluate whether each code should be accepted.

We intuitively feed $S$ into Code-Critic and output $p_\phi$ to check whether each code in $S$ is consistent with the ones in $S_h$. If not, it's rejected, otherwise it's accepted. Based on this, we can create a label sequence $M=(S \neq S_h)$. During the training stage \uppercase\expandafter{\romannumeral2}, we use the binary cross-entropy loss to optimize the Code-Critic:
\begin{equation}
\setlength\abovedisplayskip{3pt}%shrink space
\setlength\belowdisplayskip{3pt}
    \mathcal{L}_\phi=-\sum_{i=0}^N M^{(i)}\log p_\phi(S^{(i)})+(1-M^{(i)})\log(1-p_\phi(S^{(i)})).
\end{equation}

Since the Code-Critic is trained to accurately
evaluate various cases of Code-Predictor, we introduce the
sampling temperature to enhance the sampling diversity of
Code-Predictor:
\begin{equation}
    p_\theta^{(i)} = e^{\frac{p_\theta^{(i)}}{Temp}}/
    \sum_{j=0}^K e^{\frac{p_\theta^{(i)}}{Temp}},
\end{equation}
\vspace{-0.2em}
where $Temp$ is set to 2.

\subsection{Iterative Decoding Via Predictor-Critic}
\label{sec:inference}
The inference stage pipeline is illustrated in \cref{algorithm:2}. To start, the low-quality image $I_l$ is encoded by $E_L$ into $Z_l$, alongside an initial mask $M_1$. During the inference stage, the following process goes through a total of $T$ iterations. At each iteration, we obtain $Z_t$ using \cref{eq:mask}. Then, a code sequence $S$ is sampled from the Code-Predictor $G_{\theta}$. The accuracy and correlation of $S$ are evaluated using the Code-Critic $G_{\phi}$ to determine which code in $S$ should be rejected and resampled in the next iteration, as indicated by the binary mask $M_{t+1}$. The mask ratio is determined by $\lceil \gamma(r) \cdot N \rceil$. After completing $T$ iterations, we can achieve clear images with $I_{rec}=D_H(Z_c)$.

\begin{figure*}[ht] \centering
    \includegraphics[width=0.985\textwidth]{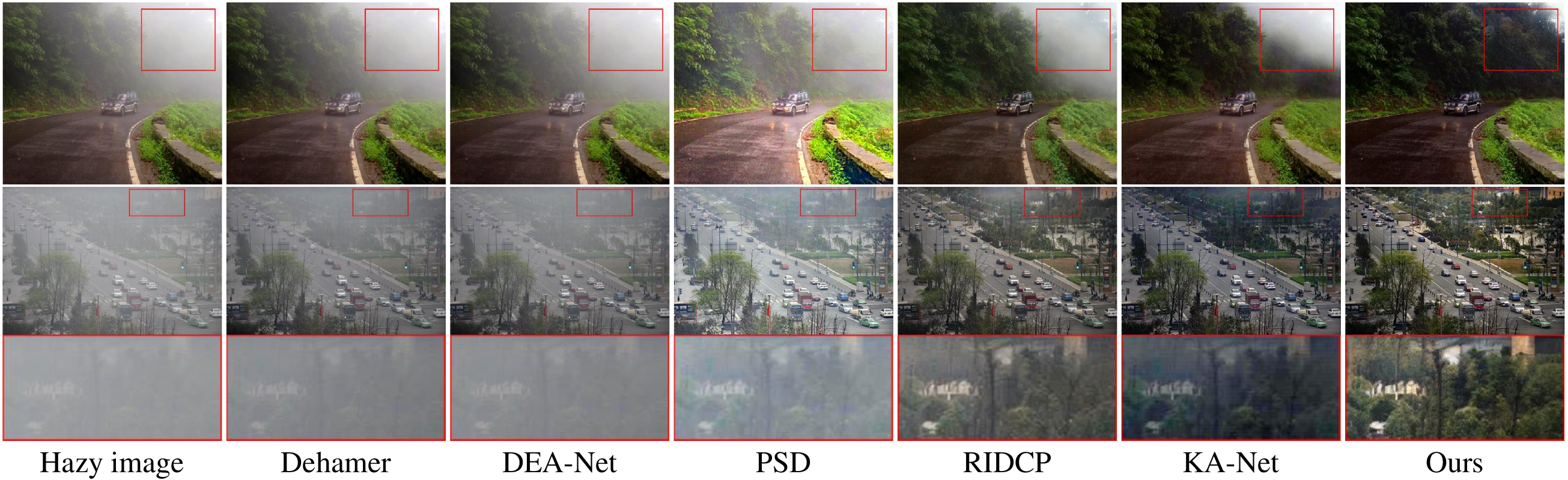}
    \caption{Visual comparison on RTTS. \textbf{Zoom in for best view}.}
    \label{fig:rtts}
\end{figure*}

\begin{figure*}[ht] \centering
    \includegraphics[width=0.985\textwidth]{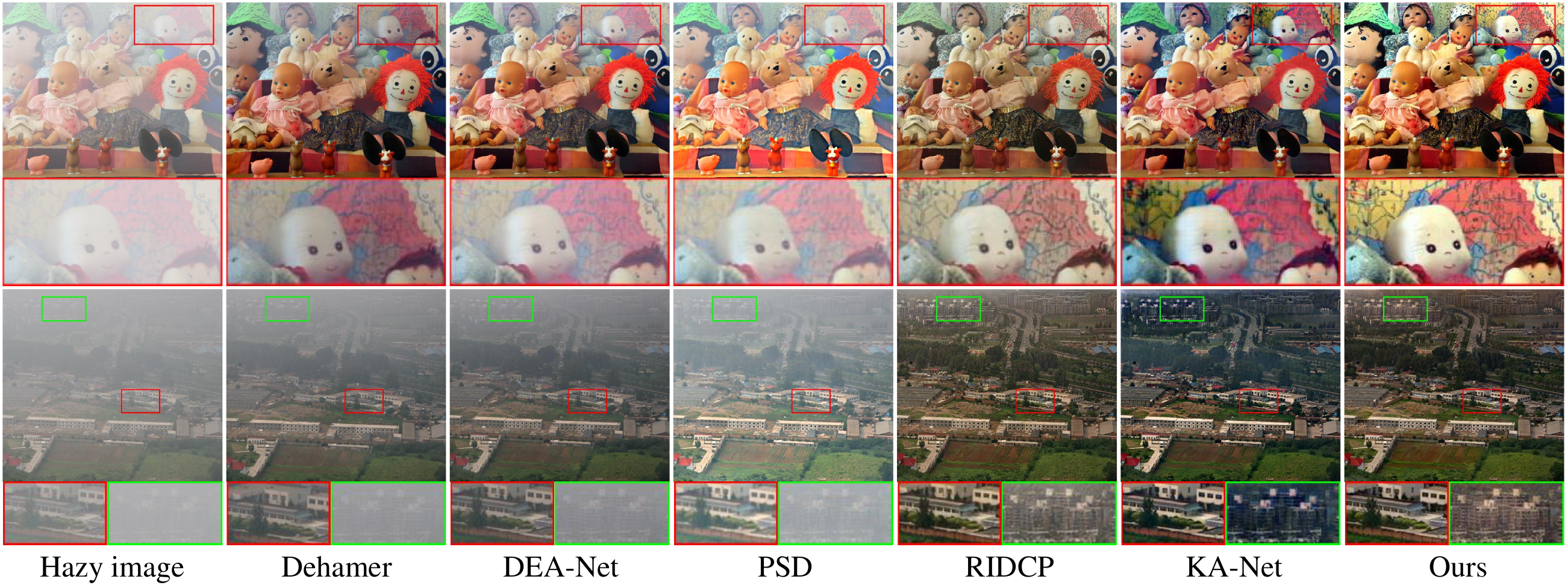}
    \caption{Visual comparison on Fattal. \textbf{Zoom in for best view}.}
    \label{fig:fattal}
\end{figure*}

\begin{figure*}[ht] \centering
    \includegraphics[width=0.985\textwidth]{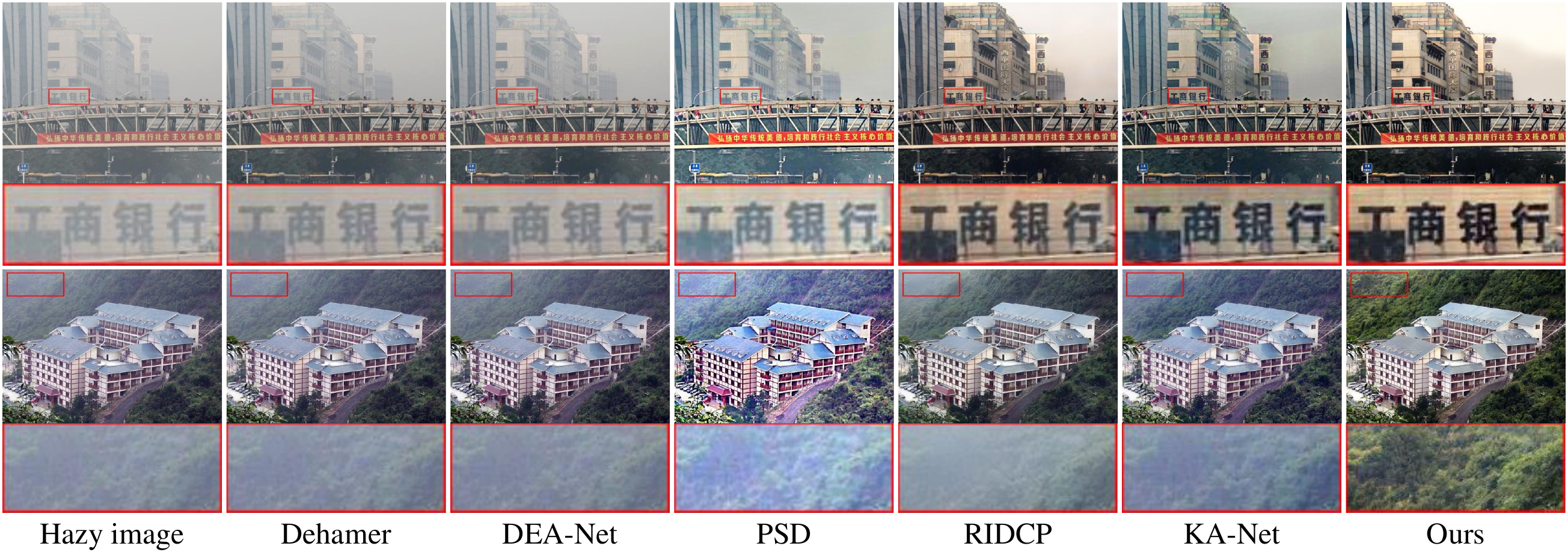}
    \caption{Visual comparison on URHI. \textbf{Zoom in for best view}.}
    \label{fig:URHI}
\end{figure*}

\begin{figure*}[ht] \centering
\includegraphics[width=0.985\textwidth]{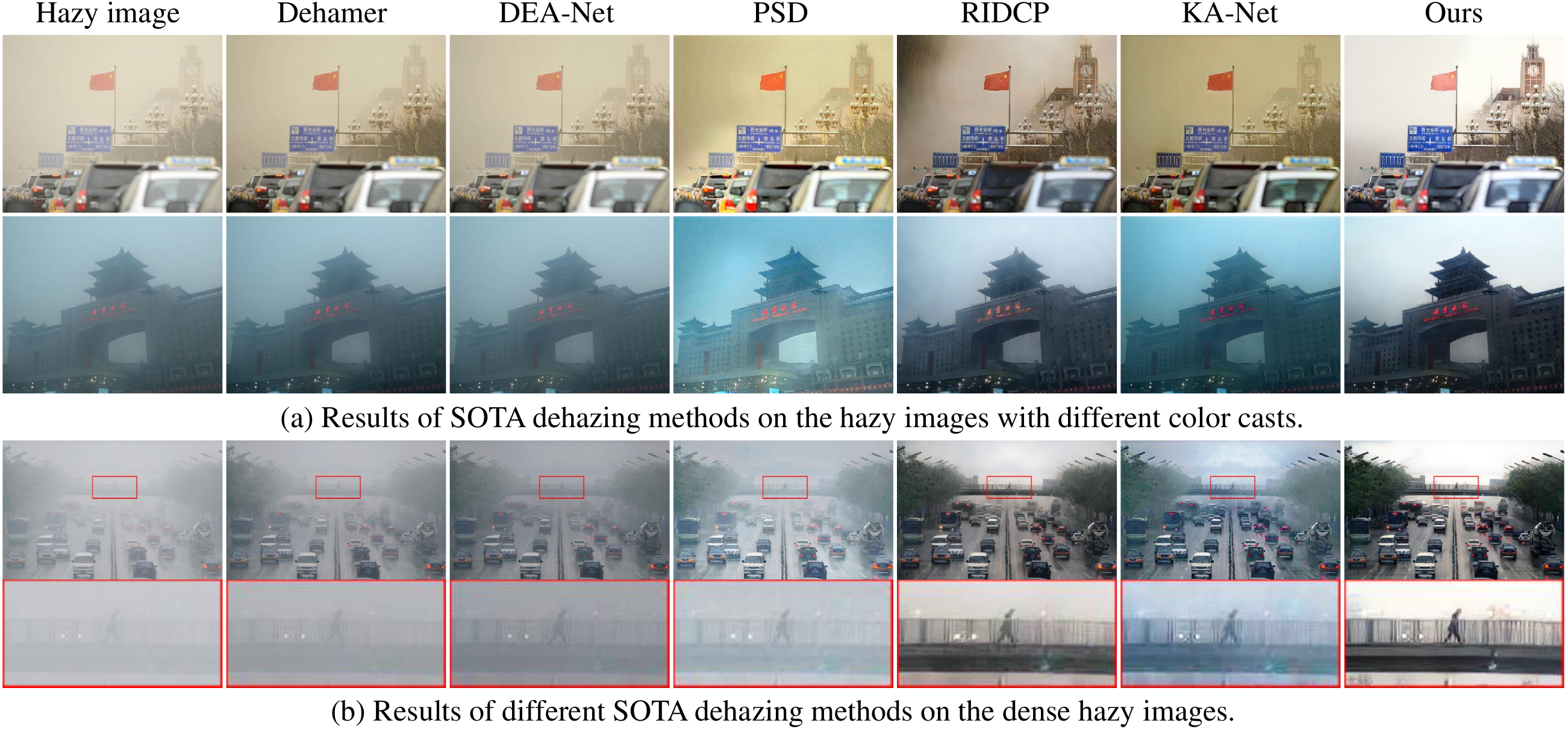}
%  \makebox[\textwidth]{\small (a) Results of different dehazing methods on the hazy images with different color casts.} 
%   \\[0.5em]
% \includegraphics[width=0.985\textwidth]{figures/mulity-case_3.pdf}   
%  \makebox[\textwidth]{\small (b) Results of different dehazing methods on the dense hazy images.} 
 
\caption{Visual comparison on challenging scenarios. \textbf{Zoom in for best view}.}
    \label{fig:challange}
\end{figure*}

\section{Experiments}

\subsection{Experimental Settings}

\smallsec{Training Datasets.} During the codebook learning phase, we follow the Real ESRGAN~\cite{Wang_Xie_Dong_Shan_2021} to generate paired data. We obtain images from DIV2K~\cite{Agustsson_Timofte_2017} and Flickr2K~\cite{Lim_Son_Kim_Nah_Lee_2017} and randomly crop and resize them into non-overlapping $512\times512$ patches. During the training phase (Stage \uppercase\expandafter{\romannumeral1} and Stage \uppercase\expandafter{\romannumeral2}), we generate paired images employing the synthetic data generation methodology proposed by RIDCP~\cite{Wu_2023_CVPR}.

\smallsec{Testing Datasets.} We qualitatively and quantitatively evaluate our model on the real-world datasets RTTS and URHI~\cite{Li_Ren_Fu_Tao_Feng_Zeng_Wang_2019}, which contain 4,322 and 4,809 images respectively, covering various scenes with different haze densities, resolutions, and degradation levels. Additionally, we further conduct comparisons on Fattal's dataset~\cite{Fattal_2014}.

\subsection{Network Architectures.} In our work, we utilize a VQGAN network similar to FeMaSR~\cite{chen2022real}. To
accommodate images of different resolutions, we employ a
4$\times$ RSTB~\cite{liang2021swinir} and a 2$\times$ RSTB as the body for the Code-Predictor and Code-Critic, followed by a linear projection layer. More details about the network can be found in the supplemental material.

Our method is implemented with PyTorch on 4 NVIDIA RTX 3090 GPUs. We randomly resize and flip the input data, and crop it into $256 \times 256$ patches for data augmentation. We use Adam with parameters of $\beta_1=0.9$, $\beta_2=0.99$, and fix the learning rate to $1\times 10^{-4}$ for all stages of training. 
We pre-train VQGAN with a batch size of 32 and adopt a batch size of 16 for Stage \uppercase\expandafter{\romannumeral1} and Stage \uppercase\expandafter{\romannumeral2}. The networks are respectively trained with 400K, 100K, and 20K iterations during the three training stages.

\subsection{Comparisons with State-of-the-art Methods}
We compare our method with several state-of-the-art dehazing approaches through both quantitative and qualitative analyses. We present more experimental results and analysis in the supplementary material.

\smallsec{Quantitative Comparison.}
Due to the difficulty in obtaining ground truth hazy images from the real world, we quantitatively analyzed our method with some commonly used no-reference image quality assessment (IQA) metrics. To better assess the results, we apply IQA metrics focusing on different aspects, such as perceptual IQA (MUSIQ~\cite{ke2021musiq}, PI~\cite{pi}, and MANIQA~\cite{yang2022maniqa}), semantic IQA (TOPIQ~\cite{chen2024topiq}), and LLMs-based IQA (CLIPIQA~\cite{wang2022exploring}, Q-Align~\cite{wu2024qalign}). All IQA metrics, except for PI, higher metric scores represent better image quality. We compare our method with the methods that have achieved outstanding results in benchmarks: MSDBN~\cite{Dong_Pan_Xiang_Hu_Zhang_Wang_Yang_2020}, Dehamer~\cite{Guo_Yan_Anwar_Cong_Ren_Li}, and DEA-Net~\cite{DEA} as well as real-world image dehazing methods: DAD~\cite{Shao_Li_Ren_Gao_Sang_2020},  PSD~\cite{Chen_Wang_Yang_Liu}, D4~\cite{Yang_Wang_Liu_Zhang_Guo_Tao}, RIDCP~\cite{Wu_2023_CVPR}, and KA-Net~\cite{KA}.

For a fair comparison, we execute the official code for all the mentioned methods and evaluate them using IQA-Pytorch~\cite{pyiqa}. As seen from the \cref{tab:1}, our method achieves first and second places in comparison to other state-of-the-art methods. The results suggest that our method performs better in dehazing capabilities, color fidelity, and image quality.
Overall, our method obtained the best results in quantitative metrics, further supporting its superiority in real-world image dehazing.

\smallsec{Qualitative Comparison.}
We conduct a qualitative comparison on the RTTS, Fattal, and URHI datasets, as shown in~\cref{fig:rtts,fig:URHI,fig:fattal}. Dehamer~\cite{Guo_Yan_Anwar_Cong_Ren_Li} and DEA-Net~\cite{DEA} show limited dehazing capabilities. PSD~\cite{Chen_Wang_Yang_Liu} is somewhat overexposed and color-shifting. RIDCP~\cite{Wu_2023_CVPR} and KA-Net~\cite{KA} demonstrate effective performance, but the image quality is subpar and they struggle with dense haze areas. In comparison, the quality of our restoration significantly outperforms other methods in both thin and dense haze regions (such as within the red and green box in the bottom image in \cref{fig:fattal}). The comparison reveals that our approach can generate high-quality, more natural, and cleaner images.

To further demonstrate the superiority of our approach, we additionally select some challenging scenarios for vision comparison. As shown in~\cref{fig:challange} (a), the captured hazy images may be color-biased due to the effects of low light, and particles such as sand, and dust. Our results can handle the color casts and have more natural colors and cleaner images. In~\cref{fig:challange} (b), our method still performs well even in the presence of dense haze with less noise compared to the suboptimal method, RIDCP.

\begin{figure}[th]
\includegraphics[width=0.47\textwidth]{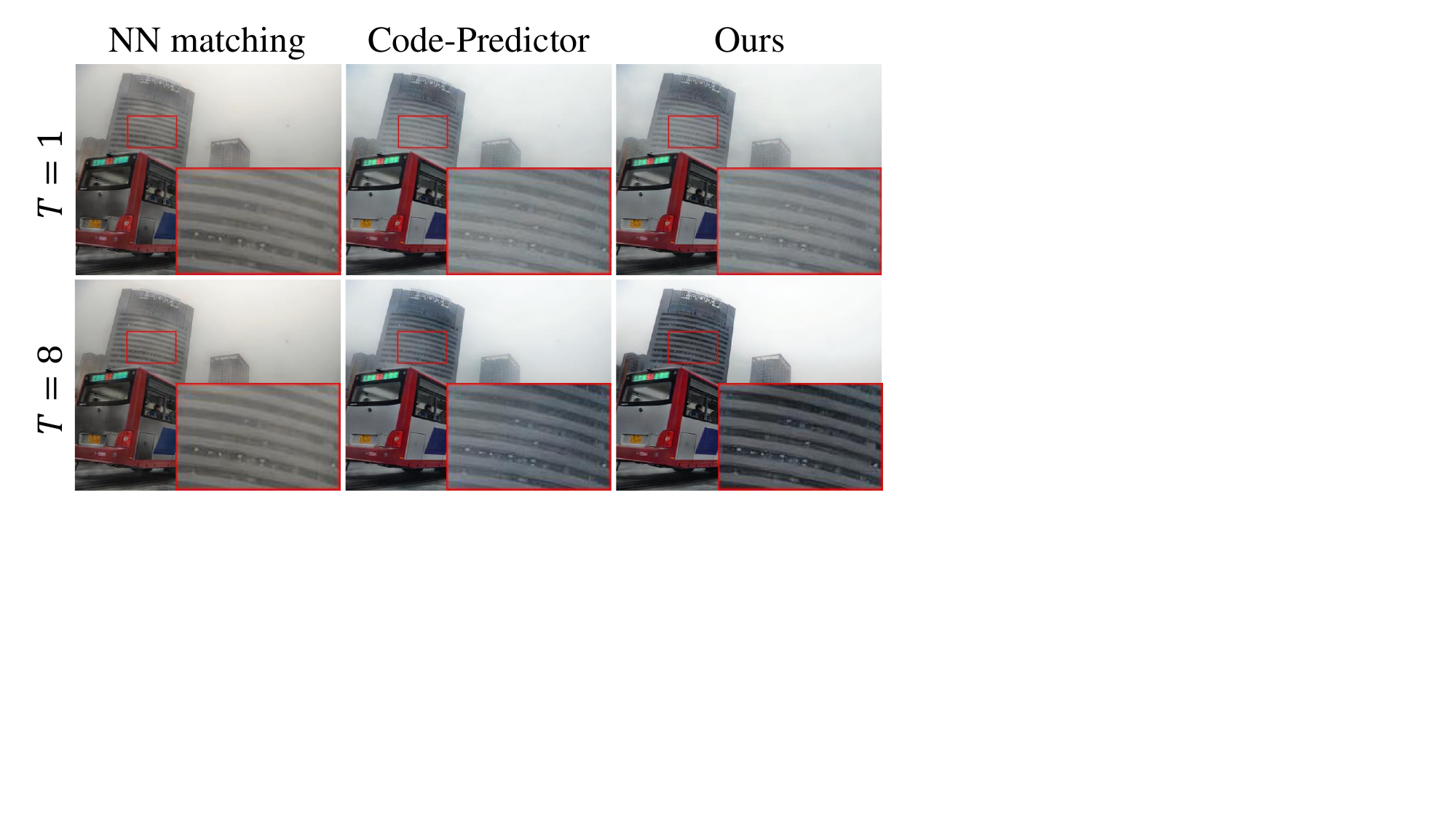}
\caption{Ablation results of the proposed Code-Predictor.
The first row displays the results with no iteration, while the second row shows the results with iteration. The first column shows the results of the NN-based code matching, the second column shows the results of Code-Predictor (without Code-Critic), and the third column shows our results. We observed that Code-Predictor allows for iterations (from $T=1$ to $T=8$) to bring about changes, and Code-Critic leads to better results.
}
    \label{fig:prediction}
\end{figure}

\begin{table}[!t]
  \centering
  \caption{Quantitative ablation analysis of the Code-Predictor on RTTS when $T = 8$. \textcolor{red}{Red} indicates the best results.}
    \resizebox{0.47\textwidth}{!}{
    \begin{tabular}{cccccc}
    \toprule
    Method & MUSIQ$\uparrow$ & PI$\downarrow$   & MANIQA$\uparrow$ & Q-Align$\uparrow$ & TOPIQ$\uparrow$ \\
    \midrule
    NN Matching & 58.19 & 3.25  & 0.303 & 3.25  & 0.458  \\
    w/o Code-Critic & 57.74 & 3.32  & 0.303 & 3.36  & 0.462  \\
    Ours  & \textcolor[rgb]{ 1,  0,  0}{59.60} & \textcolor[rgb]{ 1,  0,  0}{3.22} & \textcolor[rgb]{ 1,  0,  0}{0.327} & \textcolor[rgb]{ 1,  0,  0}{3.49 } & \textcolor[rgb]{ 1,  0,  0}{0.500 } \\
    \bottomrule
    \end{tabular}%
    }
  \label{tab:predict}%
\end{table}%

\subsection{Ablation Study and Analysis}
To validate the significance of the Code-Predictor and Code-Critic, we conducted a series of experiments focused on: (1) examining the effectiveness of iterative dehazing with Code-Predictor, and (2) investigating the enhancement of iterative dehazing through the integration of Code-Critic with Code-Predictor.
    
\smallsec{Effectiveness of Code-Predictor.} First, we compare the Code-Predictor with the nearest neighbor (NN) matching method to validate the effectiveness of this module in our network. To eliminate interference, we did not introduce the Code-Critic in this ablation experiment. In order to apply iteration in the model based on nearest neighbor matching, we use the distance from token mapping to the codebook as the criterion for whether the generated code will be retained in the next iteration. As shown in \cref{fig:prediction}, when using NN matching, increasing the number of iterations does not lead to better results (see the first column). It is obvious that during the training phase, what we learn is independent matching for each token without considering the relationships between tokens, so iteration does not change the code. However, Code-Predictor takes $Z_l$ as a condition, which uses the high-quality code obtained from the previous iteration to guide the next iteration.

Compared to the NN matching method, our Code-Predictor can select more likely codes during the iteration process, which is crucial for the iterative decoding prediction. \cref{tab:predict} also verifies this conclusion from the point of view of quantitative analysis. However, without the Code-Critic, Code-Predictor cannot effectively determine which codes to keep in each iteration, resulting in limited improvements. We further discuss the role of Code-Critic below.

\begin{figure}[!t] \centering
    \includegraphics[width=0.47\textwidth]{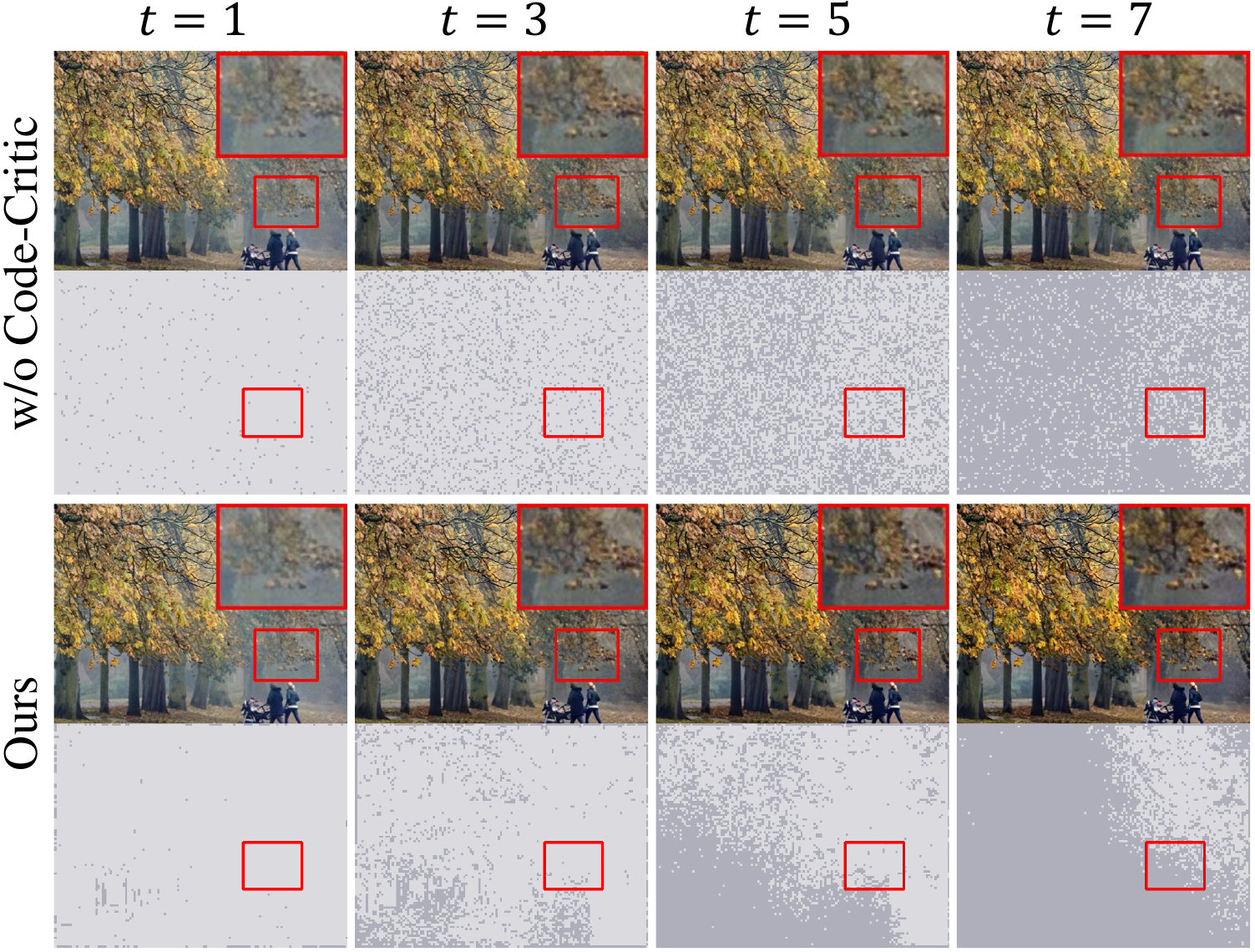}
    \caption{Ablation results of the proposed Code-Critic. The images from left to right display the results at different $t$ When $T = 8$. The top image in each pair represents the results from sequence $S$, while the bottom image represents the mask map after $S$ has been evaluated by Code-Critic. Without Code-Critic, the mask map tends to be random. Our method’s mask map can first retain the code in thin haze areas, resulting in better outcomes.}
    \label{fig:critic}
\end{figure}

\smallsec{Effectiveness of Code-Critic.}
To further discuss the necessity of Code-Critic, we compare the sampling method based on code confidence and that based on Code-Critic. As shown in~\cref{algorithm:2}, the results are shown in \cref{fig:prediction,fig:critic}. It is evident that with the inclusion of the Code-Critic, the process of code selection exhibits a trend from close to distant, from simple to challenging. Moreover, significant improvements in image optimization are observed during iterations. Such a design leads to much cleaner and sharper results in comparison to the absence of the Code-Critic. Additionally,~\cref{tab:predict} further demonstrates the findings.

\section{Conclusion}
Inspired by the natural phenomenon that the degradation of haze images is closely related to the density of haze and scene depth, we propose a new method for real-world iterative dehazing. Our iterative process is different from the one-shot-based method. The one-shot-based method cannot effectively handle thin haze areas. In contrast, our process can first identify faint haze and restore regions with low difficulty. Then, it uses these restored parts as guidance to predict haze density and restore regions with higher difficulty. Extensive experiments demonstrate the superiority of our approach.

\smallsec{Acknowledgements.}
This work was supported in part by the National Natural Science Foundation of China (62306153, U23B2011, 62176130), the Fundamental Research Funds for the Central Universities (Nankai University, 070-63243143), the Natural Science Foundation of Tianjin, China (24JCJQJC00020), the Shenzhen Science and Technology Program (JCYJ20240813114237048) and the Key R\&D Program of Zhejiang (2024SSYS0091). The computational devices of this work are supported by the Supercomputing Center of Nankai University (NKSC).

\clearpage
\maketitlesupplementary
\appendix
\begin{abstract}
This supplementary material presents the network architectures, objective functions, further ablation experiments, and more visual results. Specifically, in ablation experiments, we include ablation results on the real-world URHI dataset to validate the effectiveness of the Code-Critic in~\cref{sup-critic}. Besides, throughout the paper, we typically set $T=8$. To justify this setting, we discuss the effect of the number of iterations in \cref{sup-iteration}. 
\end{abstract}

\section{Network Architectures}
The structural details of IPC-Dehaze are shown in \cref{tab:arch}. In VQGAN, we use a network structure similar to FeMaSR~\cite{chen2022real} and set the codebook size $K$ to 1024 and embedding dim to 256. To achieve code matching and code evaluation at different resolutions, we use RSTB~\cite{liang2021swinir} as the backbone and add a linear projection layer.
\begin{table}[thbp]
  \centering
  \small
     \resizebox{0.47\textwidth}{!}{
    \begin{tabular}{c|c|c}
    \toprule
    \toprule
    Layers & Configuration & Output Size \\
    \midrule
    \textcolor[rgb]{ 0,  0,  1}{Conv\_in} & \textcolor[rgb]{ 0,  0,  1}{$c_{in}=3 \ c_{out}=64 \ ksz=4$}& \textcolor[rgb]{ 0,  0,  1}{$(h, w, 64)$}\\
    \midrule
    \textcolor[rgb]{ 0,  0,  1}{Block1} & \textcolor[rgb]{ 0,  0,  1}{$c_{in}=64\ c_{out}=128\ ksz=3\ stride=2$} & \textcolor[rgb]{ 0,  0,  1}{$(h/2, w/2, 128)$}\\
    \midrule
    \textcolor[rgb]{ 0,  0,  1}{Block2} & \textcolor[rgb]{ 0,  0,  1}{$c_{in}=128\ c_{out}=256\ ksz=3\ stride=2$} & \textcolor[rgb]{ 0,  0,  1}{$(h/4, w/4, 256)$}\\
    \midrule
    \textcolor[rgb]{ 0,  0,  1}{Before\_quant} & \textcolor[rgb]{ 0,  0,  1}{$c_{in}=256\ c_{out}=256\ ksz=1\ stride=1$} & \textcolor[rgb]{ 0,  0,  1}{$(h/4, w/4, 256)$}\\
    \midrule
    \textcolor[rgb]{ 0,  .69,  .314}{RSTB\_block1} & \textcolor[rgb]{ 0,  .69,  .314}{$\left[ \begin{array}{l}
{\rm{  }}ws{\rm{   }} = {\rm{    }}8\\
{\rm{   }}d{\rm{    }} = {\rm{ }}256\\
head{\rm{  }} = {\rm{   }}8\\
depth{\rm{ }} = {\rm{   }}6
\end{array} \right] \times 4$} & \textcolor[rgb]{ 0,  .69,  .314}{$(h/4, w/4, 256)$}\\
    \midrule
    \textcolor[rgb]{ 0,  .69,  .314}{Norm1} & \textcolor[rgb]{ 0,  .69,  .314}{$d$=256} & \textcolor[rgb]{ 0,  .69,  .314}{$(h/4, w/4, 256)$}\\
    \midrule
    \textcolor[rgb]{ 0,  .69,  .314}{Linear1} & \textcolor[rgb]{ 0,  .69,  .314}{$f_{in}=256\ f_{out}=1024$} & \textcolor[rgb]{ 0,  .69,  .314}{$(h/4, w/4, 1024)$}\\
    \midrule
    \textcolor[rgb]{ .776,  .349,  .067}{RSTB\_block2} & \textcolor[rgb]{.776,.349,.067}{$\left[ \begin{array}{l}
{\rm{  }}ws{\rm{   }} = {\rm{    }}8\\
{\rm{   }}d{\rm{    }} = {\rm{ }}256\\
head{\rm{  }} = {\rm{   }}8\\
depth{\rm{ }} = {\rm{   }}6
\end{array} \right] \times 2$}& \textcolor[rgb]{ .776,  .349,  .067}{$(h/4, w/4, 256)$}\\
    \midrule
    \textcolor[rgb]{ .776,  .349,  .067}{Norm2} & \textcolor[rgb]{ .776,  .349,  .067}{$d$=256} & \textcolor[rgb]{ .776,  .349,  .067}{$(h/4, w/4, 256)$}\\
    \midrule
    \textcolor[rgb]{ .776,  .349,  .067}{Linear2} & \textcolor[rgb]{ .776,  .349,  .067}{$f_{in}=256\ f_{out}=1$} & \textcolor[rgb]{ .776,  .349,  .067}{$(h/4, w/4, 1)$}\\
    \midrule
    \textcolor[rgb]{ 0,  0,  1}{Codebook} & \textcolor[rgb]{ 0,  0,  1}{$K=1024\ d=256$} & \textcolor[rgb]{ 0,  0,  1}{$(h/4, w/4, 256)$}\\
    \midrule
    \textcolor[rgb]{ 0,  0,  1}{After\_quant} & \textcolor[rgb]{ 0,  0,  1}{$c_{in}=256\ c_{out}=256\ ksz=3\ stride=1$} & \textcolor[rgb]{ 0,  0,  1}{$(h/4, w/4, 256)$}\\

    \midrule
    \textcolor[rgb]{ 0,  0,  1}{Upsample} & \textcolor[rgb]{ 0,  0,  1}{ratio=2} & \textcolor[rgb]{ 0,  0,  1}{$(h/2, w/2, 256)$}\\
    \midrule
    \textcolor[rgb]{ 0,  0,  1}{Block3}& \textcolor[rgb]{ 0,  0,  1}{$c_{in}=256\ c_{out}=128\ ksz=3\ stride=1$} & \textcolor[rgb]{ 0,  0,  1}{$(h/2, w/2, 128)$}\\
    \midrule
    \textcolor[rgb]{ .749,  .561,  0}{SFT\_block1} & \textcolor[rgb]{ .749,  .561,  0}{$c_{in}=128\ c_{out}=128\ ksz=3\ stride=1$} & \textcolor[rgb]{ .749,  .561,  0}{$(h/2, w/2, 128)$}\\
    \midrule
    \textcolor[rgb]{ 0,  0,  1}{Upsample} & \textcolor[rgb]{ 0,  0,  1}{ratio=2} & \textcolor[rgb]{ 0,  0,  1}{$(h, w, 128)$}\\
    \midrule
    \textcolor[rgb]{ 0,  0,  1}{Block4}& \textcolor[rgb]{ 0,  0,  1}{$c_{in}=128\ c_{out}=64\ ksz=3\ stride=1$} & \textcolor[rgb]{ 0,  0,  1}{$(h, w, 64)$}\\
    \midrule
    \textcolor[rgb]{ .749,  .561,  0}{SFT\_block2} & \textcolor[rgb]{ .749,  .561,  0}{$c_{in}=64\ c_{out}=64\ ksz=3\ stride=1$} & \textcolor[rgb]{ .749,  .561,  0}{$(h, w, 64)$}\\
    \midrule
    \textcolor[rgb]{ 0,  0,  1}{Conv\_{out}} & \textcolor[rgb]{ 0,  0,  1}{$c_{in}=64\ c_{out}=3\ ksz=3\ stride=1$} & \textcolor[rgb]{ 0,  0,  1}{$(h, w, 3)$}\\
    \bottomrule
    \bottomrule
    \end{tabular}%
    }
  \caption{Architecture details of the IPC-Dehaze. \textcolor{blue}{Blue}, \textcolor[rgb]{ 0,  .69,  .314}{green},  \textcolor[rgb]{ .776,  .349,  .067}{brown}, and  \textcolor[rgb]{ .749,  .561,  0}{yellow} represent the layers of VQGAN, Code-Predictor, Code-Critic, and SFT, respectively. $c_{in}$, $c_{out}$, and $ksz$ are the input channel, output channel, and kernel size, respectively. The $ws$ is the window size and $d$ is the embedding dim. $f_{in}$ is the number of input features and $f_{out}$ is the number of output features. The input of the network is an RGB image  $\in \mathbb{R}^{h\times w\times 3}$.}
    \label{tab:arch}%
\end{table}%

\section{Objective Functions}
\subsection{Pretrain: VQGAN Training}
Following VQGAN, we adopt pixel-level reconstruction loss $\mathcal{L}_1$ and code-level loss $\mathcal{L}_{code}$ to train the Encoder $E_H$, Decoder $D_H$, and Codebook $C$:
\begin{equation}
    \mathcal{L}_{1} = \Vert I_h - I_{rec} \Vert_1,
\end{equation}
\begin{equation}
    \begin{aligned}
        \mathcal{L}_{code} &= \Vert sg(z_c)-Z_h^{(i,j)} \Vert_2^2 + \beta \Vert z_c - sg(Z_h^{(i,j)}) \Vert_2^2 \\
        &\hspace{5em} +\lambda_{g}\Vert\text{CONV}(Z_h^{(i,j)})-\Phi(I_h)\Vert_2^2,
    \end{aligned}
\end{equation}
where $sg(\cdot)$ is the stop-gradient operation, $\beta=0.25$ and $\lambda_{g}=0.1$ in this training. The \text{CONV}($\cdot$) and $\Phi$($\cdot$) denote a convolution layer and a pre-trained VGG19~\cite{vgg19}, respectively.

To restore better texture, we use perceptual loss $\mathcal{L}_{per}$~\cite{perceploss}, and adversarial loss $\mathcal{L}_{adv}$~\cite{ganloss} as part of the loss function:
\begin{equation}
    \mathcal{L}_{per}=\Vert \Phi(I_h)-\Phi(I_{rec}) \Vert_1,
\end{equation}
\begin{equation}
    \mathcal{L}_{adv}= \log D(I_h)+\log(1-D(I_{rec})).
\end{equation}

In the pre-training stage, the loss is expressed as:
\begin{align}
    \mathcal{L}_{VQGAN} &= \mathcal{L}_{1} +  \mathcal{L}_{code}+\mathcal{L}_{per} + \lambda_{adv}\mathcal{L}_{adv},
\end{align}
where $\lambda_{adv}=0.1$.
\subsection{Stage \uppercase\expandafter{\romannumeral1}: Code-Predictor Training
}
In this training stage, we use $\mathcal{L}_{1}$, $\mathcal{L}_{per}$, and $\mathcal{L}_{adv}$ to train the Encoder $E_L$. In addition, we use the cross-entropy loss $\mathcal{L}_\theta$ to train the Code-Predictor. The total loss is expressed as:
\begin{equation}
    \mathcal{L}_\theta = -\sum_{i=0}^N S_h^{(i)} \log p_\theta(S^{(i)}|Z_t),
\end{equation}
\begin{equation}
    \mathcal{L}_{total} = \mathcal{L}_{1} + \mathcal{L}_{per} +  \lambda_{adv}\mathcal{L}_{adv} + \mathcal{L}_{\theta},
\end{equation}
where $\lambda_{adv}=0.1$, $S_h$ is the code sequence from the clean image, and $S$ is the code sequence predicted by Code-Predictor. $Z_t$ denotes the fused tokens and $p_\theta$ denotes the output distribution of Code-Predictor.
\subsection{Stage \uppercase\expandafter{\romannumeral2}: Code-Critic Training}

In the training stage \uppercase\expandafter{\romannumeral2}, we keep all other modules fixed, exclusively train the Code-Critic, and utilize only binary cross-entropy loss:
\begin{equation}
    \mathcal{L}_\phi=-\sum_{i=0}^N M^{(i)}\log p_\phi(S^{(i)})+(1-M^{(i)})\log(1-p_\phi(S^{(i)})),
\end{equation}
where $M = (S_h \neq S)$ and $p_\phi$ denotes the output of Code-Critic.

Since the Code-Critic module is only trained to make an accurate evaluation of Code-Predictor's diverse cases, we introduce the sampling temperature when Code-Predictor samples the code sequence $S$. The partial code is shown in ~\cref{code}.

\lstset{
    language=Python,
    basicstyle=\footnotesize\ttfamily,
    numbers=left,
    numbersep=5pt,
    frame=lines,
    framerule=0.5pt
}
\begin{figure}[!ht]
\begin{lstlisting}
# Fuse the hq_feats and lq_feats with mask    
input_feats=hq_feats*~mask+lq_feats*mask
# Get the logits with [b, h*w, K].
logits = net_predictor.transformer(input_feats)
# Add sampling temperature
logits/=Tem
probs = F.softmax(logits, -1)
# Samples the id.
sampled_ids = torch.multinomial(probs, 1)
# Evaluate the sampled_ids
masked_logits = net_critic(sampled_ids,h,w)
\end{lstlisting}
\caption{Partial code for Code-Critic Trainin.}\label{code}
\end{figure}

\section{Ablation Experiments}

\subsection{Effectiveness of Code-Critic}
\label{sup-critic}
To further discuss the necessity of Code-Critic, we compare the sampling method based on code confidence and that based on Code-Critic. We 
 conduct quantitative experience on two real-world datasets RTTS and URHI~\cite{Li_Ren_Fu_Tao_Feng_Zeng_Wang_2019}, which both contain over 4,000 images. We present the results in~\cref{sup-tab1}.

% 完整的表格
\begin{table}[htbp]
  \centering
  \caption{Quantitative ablation analysis of the Code-Predictor. \textcolor{red}{Red} indicates the best results. In this experiment, we set $T = 8$.}
   \makebox[0.48\textwidth]{\small (a) Results on RTTS.} 
    \vspace{0.5em}
    \resizebox{0.48\textwidth}{!}{

    \begin{tabular}{ccccccc}
    \toprule
    Method & MUSIQ$\uparrow$ & PI$\downarrow$   & MANIQA$\uparrow$  &CLIPIQA$\uparrow$& Q-Align$\uparrow$ & TOPIQ$\uparrow$ \\
    \midrule
    NN Matching Based & 58.19 & 3.25  & 0.303  &0.391& 3.25  & 0.458  \\
    Ours (w/o Code-Critic) & 57.74 & 3.32  & 0.303  &0.412& 3.36  & 0.462  \\
    Ours  & \textcolor{red}{59.60} & \textcolor{red}{3.22} & \textcolor{red}{0.327}  &\textcolor{red}{0.44}& \textcolor{red}{3.49 } & \textcolor{red}{0.500 } \\
    \bottomrule
    \end{tabular}%
    }
  \makebox[0.48\textwidth]{\small (b) Results on URHI.} 
    \resizebox{0.48\textwidth}{!}{
     \begin{tabular}{ccccc}
    \toprule
    Method & MUSIQ$\uparrow$ & PI$\downarrow$   & MANIQA$\uparrow$  &Q-Align$\uparrow$ 
\\
    \midrule
    NN Matching Based & 62.06& \textcolor{red}{3.01}& 0.350&3.48
\\
    Ours (w/o Code-Critic) & 60.51& 3.13& 0.343&3.55
\\
    Ours  & \textcolor{red}{62.50}& 3.08& \textcolor{red}{0.364}&\textcolor{red}{3.70}\\
    \bottomrule
    \end{tabular}%
    }
\label{sup-tab1}

\end{table}

\subsection{Iteration Number}
\label{sup-iteration}
We investigate the impact of varying the iteration number $T$ on inference performance, as summarized in \cref{tab:number}. To balance performance and efficiency, we set $T=8$ as the default. As shown in \cref{tab:number}, reducing $T$ slightly degrades performance but still outperforms other methods. For faster inference, a smaller $T$ can be selected without the need for network retraining.

\begin{table}[htbp]
  \centering
  \caption{Quantitative analysis of the different iteration numbers. \textcolor{red}{Red} indicates the best results.}
   \resizebox{0.48\textwidth}{!}{
    \begin{tabular}{ccccccc}
    \toprule
 $T$ & MUSIQ$\uparrow$ & PI$\downarrow$   & MANIQA$\uparrow$  &CLIPIQA$\uparrow$& Q-Align$\uparrow$ & TOPIQ$\uparrow$ \\
 \midrule
    3     & 58.40  & 3.33  & 0.314 & 0.425 & 3.43  & 0.478 \\
    4     & 58.71 & 3.31  & 0.318 & 0.431 & 3.45  & 0.484 \\
    6     & 58.97 & 3.30  & 0.321 & 0.437 & 3.47  & 0.490  \\
    8     &     \textcolor{red}{59.60 } & \textcolor{red}{3.22} & \textcolor{red}{0.327} & \textcolor{red}{0.444} & \textcolor{red}{3.489} & \textcolor{red}{0.500 } \\
    10    & 59.23 & 3.28  & 0.325 & 0.442 & 3.488 & 0.496 \\
    \bottomrule
    \end{tabular}%
    }
  \label{tab:number}%
\end{table}%

\section{Visual Results}
\subsection{Visualisation of Iterative Decoding}
\begin{figure*}[htbp] \centering
\includegraphics[width=0.8\textwidth]{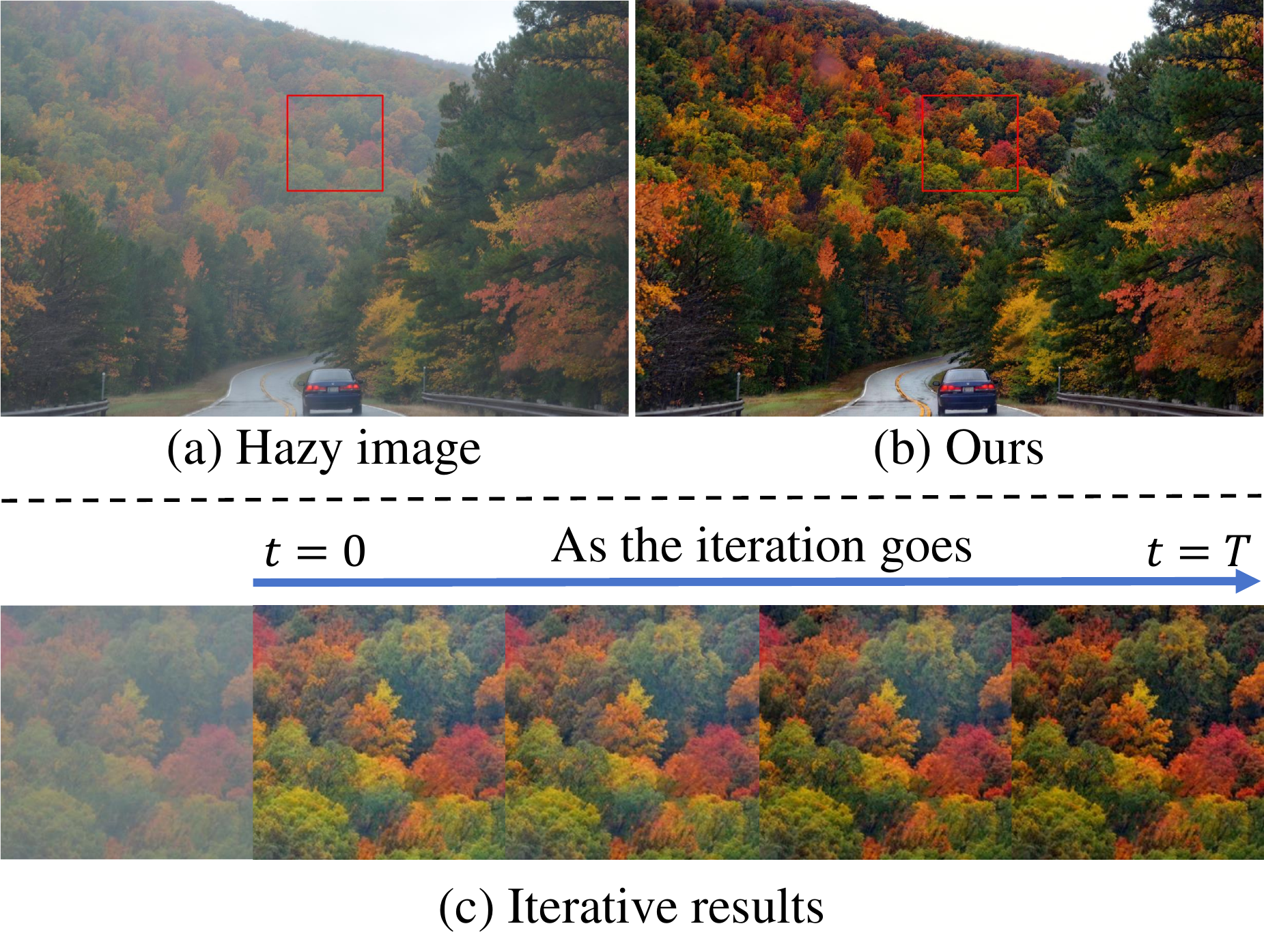}
\includegraphics[width=0.8\textwidth]{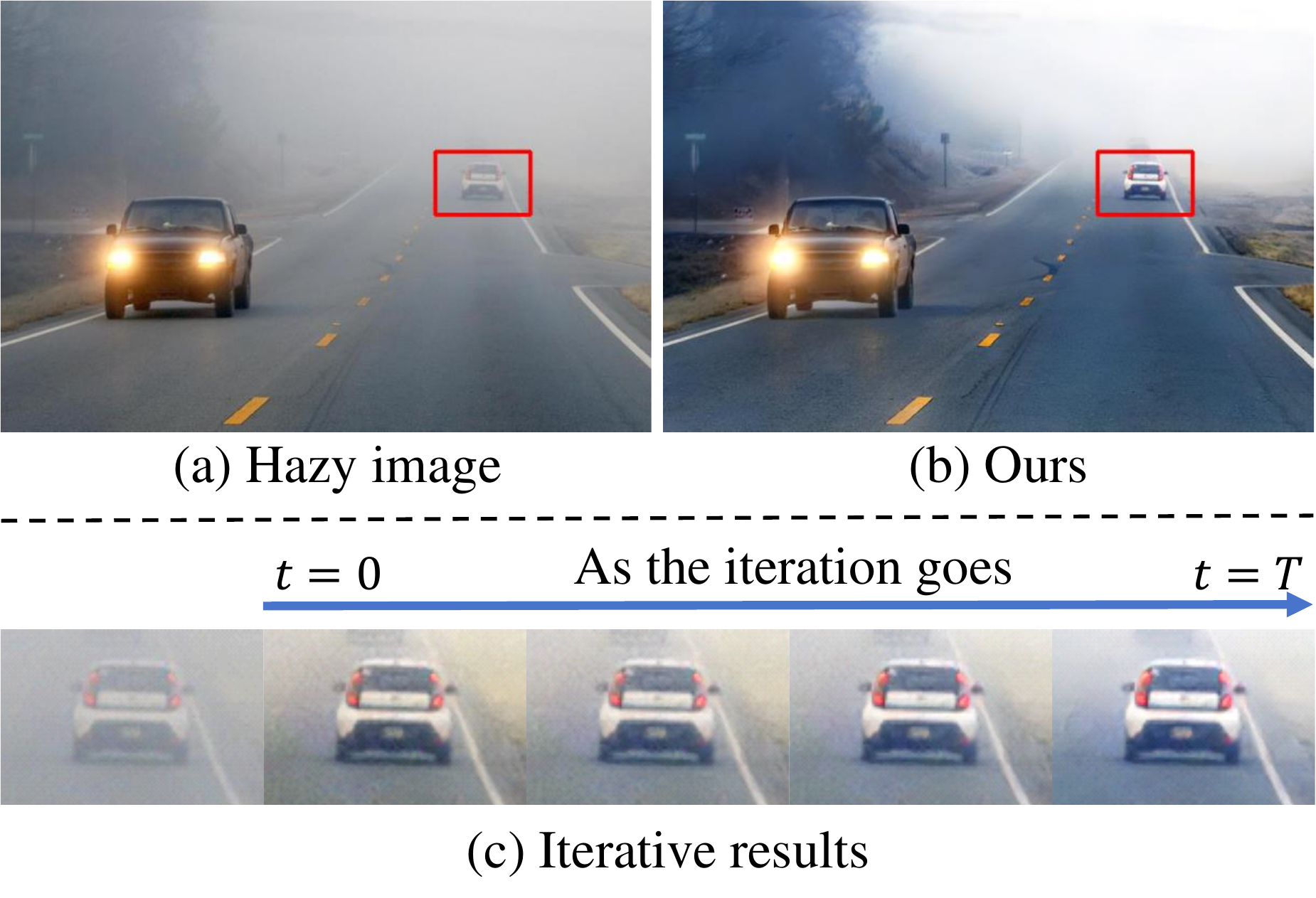}
\caption{The top-left shows the input image, while the top-right displays our result. Below, the images from left to right depict the intermediate results of the iterative decoding process when $T=8$.}
    \label{fig:iter}
\end{figure*}

\cref{fig:iter} visualizes the iterative process, and it can be seen that during the iterative process, our results get better and better in terms of dehazing, image quality, and visual effects.

\subsection{More Visual Comparisons}
\begin{figure*}[ht]
    \centering
\includegraphics[width=0.95\textwidth]{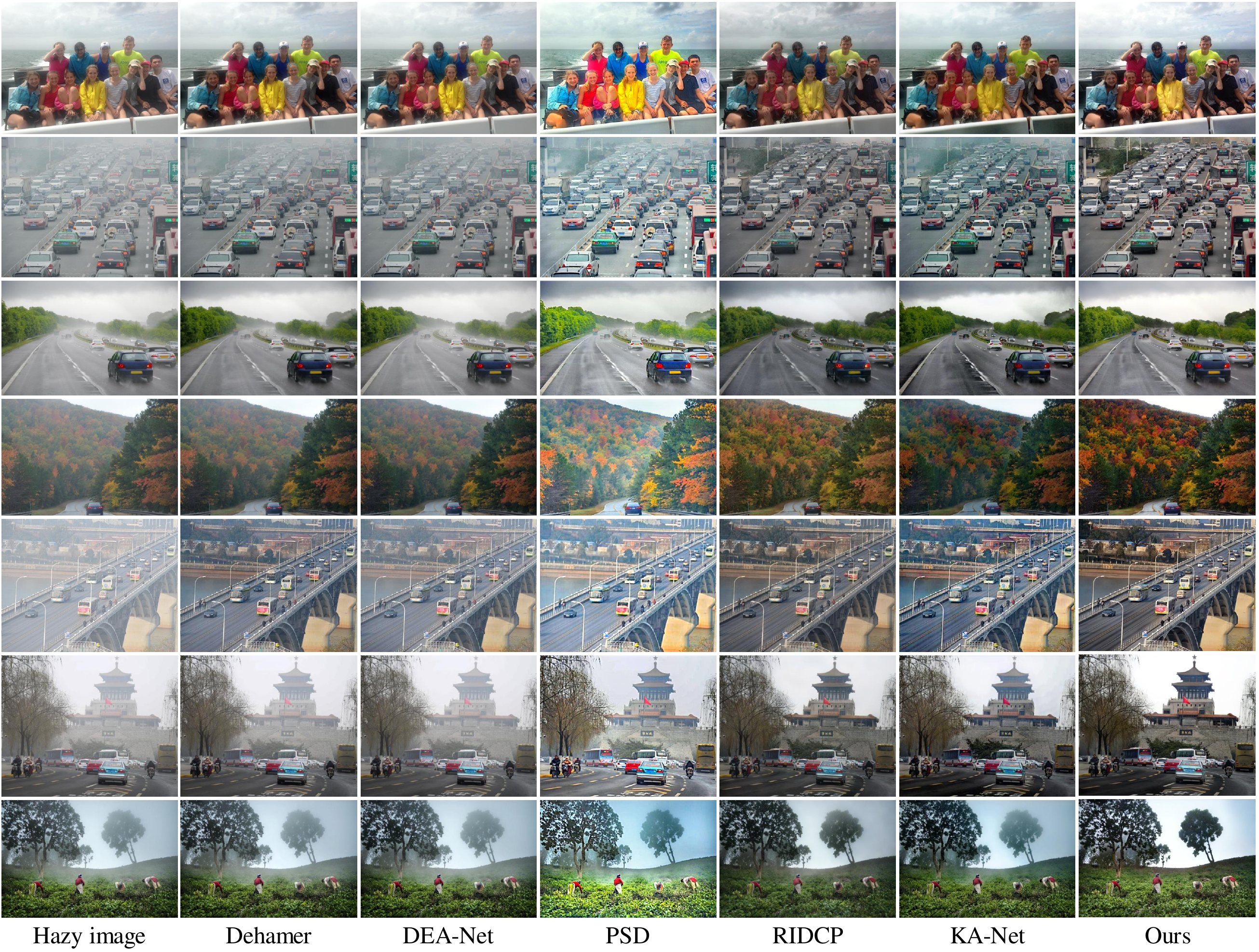}

    \caption{Visual comparison on RTTS. \textbf{Zoom in for best view}.}
    \label{fig:suprtts}
\end{figure*}

\begin{figure*}[ht]
    \centering
    \includegraphics[width=0.95\textwidth]{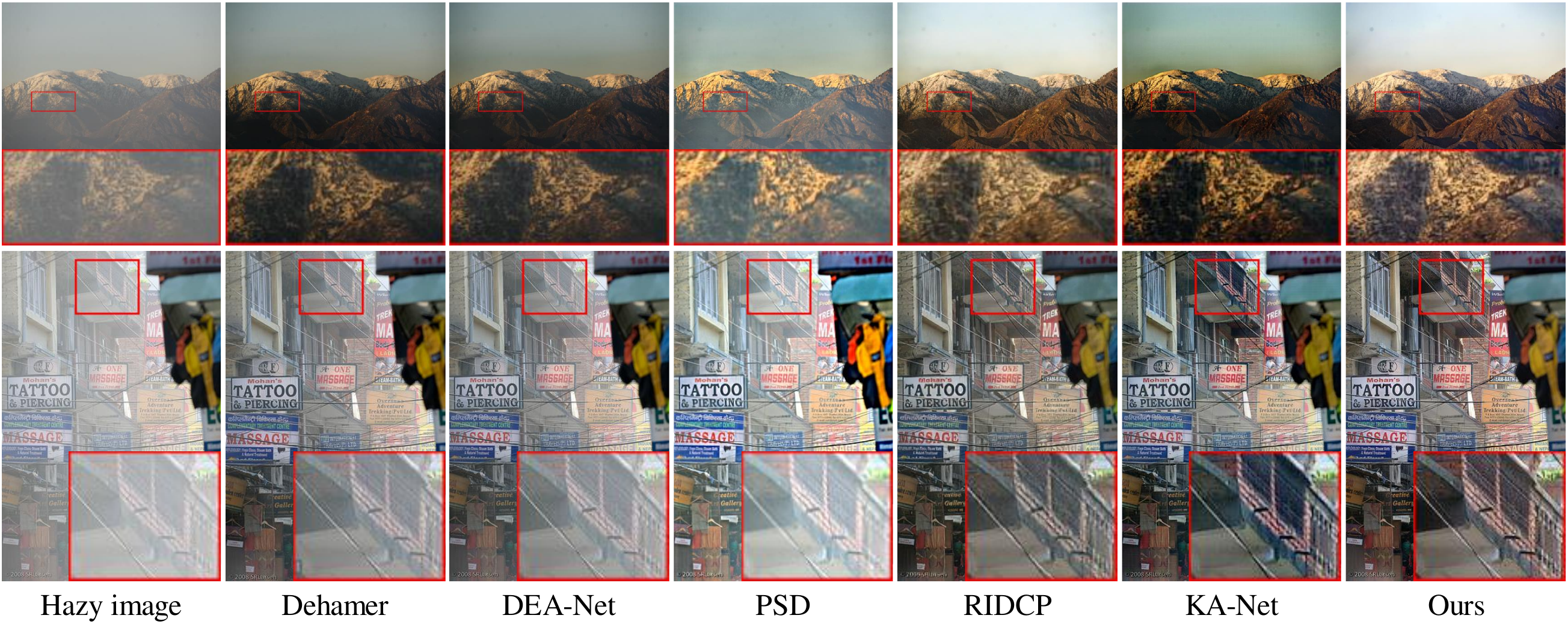}
    \caption{Visual comparison on Fattal. \textbf{Zoom in for best view}.}
    \label{fig:supfattal}
\end{figure*}

\begin{figure*}[ht]
    \centering
    \includegraphics[width=0.95\textwidth]{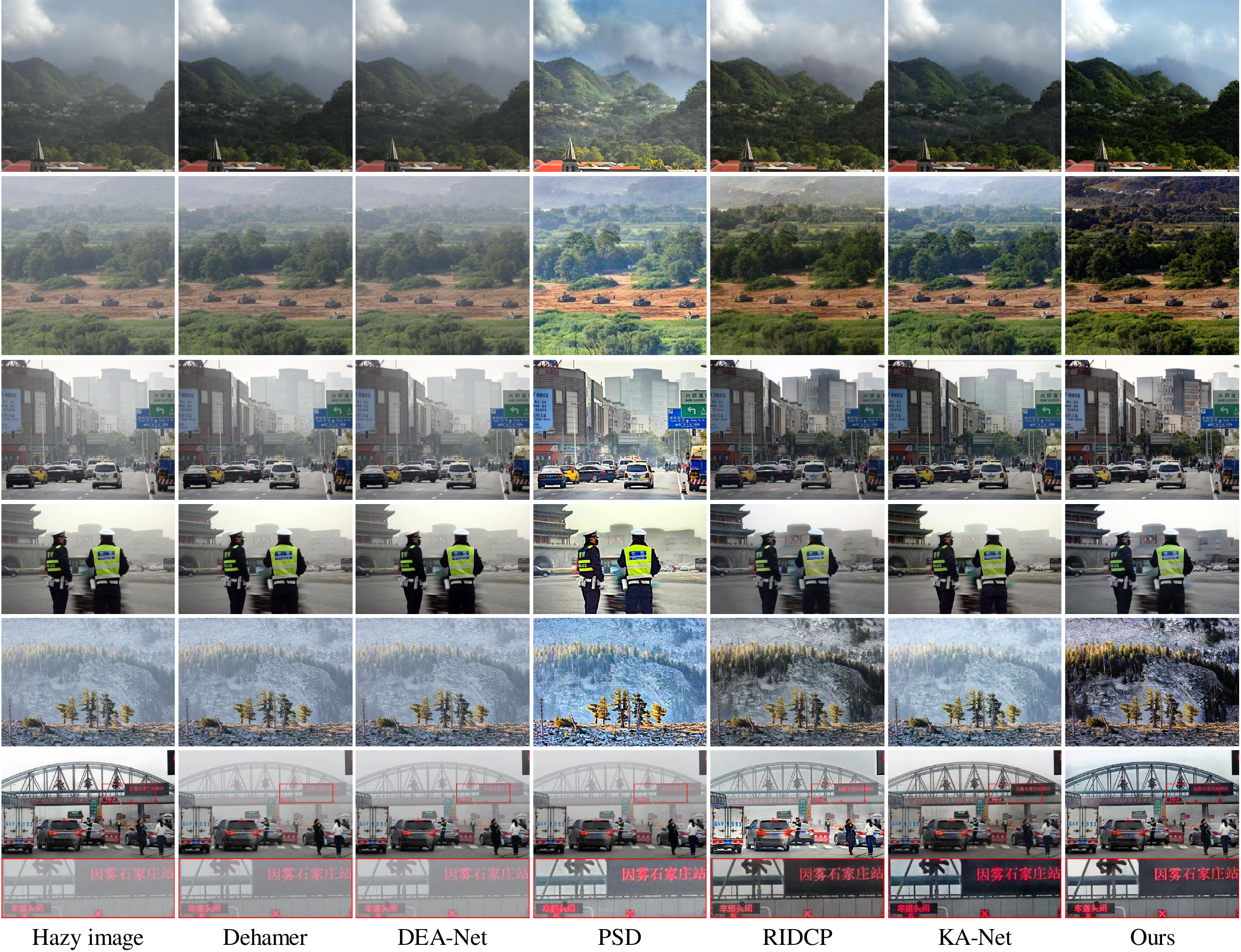}
    \caption{Visual comparison on URHI. \textbf{Zoom in for best view}.}
    \label{fig:supurhi}
\end{figure*}

\cref{fig:suprtts,fig:supfattal,fig:supurhi} show visual comparisons on the RTTS, URHI~\cite{Li_Ren_Fu_Tao_Feng_Zeng_Wang_2019} and Fattal~\cite{Fattal_2014} datasets. We compare our method with the methods that have achieved outstanding results in benchmarks: MSDBN~\cite{Dong_Pan_Xiang_Hu_Zhang_Wang_Yang_2020}, Dehamer~\cite{Guo_Yan_Anwar_Cong_Ren_Li}, and DEA-Net~\cite{DEA} as well as real-world image dehazing methods: DAD~\cite{Shao_Li_Ren_Gao_Sang_2020},  PSD~\cite{Chen_Wang_Yang_Liu}, D4~\cite{Yang_Wang_Liu_Zhang_Guo_Tao}, RIDCP~\cite{Wu_2023_CVPR}, and KA-Net~\cite{KA}. The results show that our method can achieve more natural, high-quality images, especially in dense hazy areas, which is a significant improvement compared to other methods.

\subsection{Failure Case}
\label{sup-challange}
\begin{figure*}[htbp]
    \centering
    \includegraphics[width=0.95\textwidth]{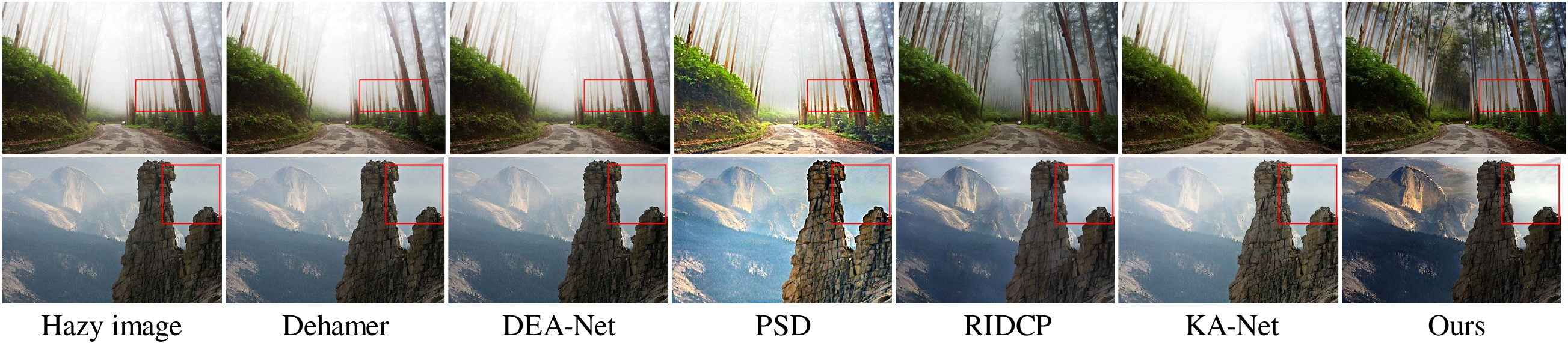}
    \caption{Failure Case. As shown in the red boxes, depth-discontinuous regions present significant challenges for existing dehazing methods. While our method performs well in the depth-continuous regions outside the red box, it struggles inside the red box, where trees and rocks disrupt depth continuity, rendering our method as ineffective as other approaches.}
    \label{fig:fail}
\end{figure*}

In this part, we discuss the failure case of our method. To be specific, our method exhibits a progressive approach to dehazing, processing denser haze regions first and gradually moving to thinner regions during iterative dehazing. This gives us a distinct advantage in handling depth-continuous scenes, as high-quality codes from earlier iterations can serve as cues for the Code-Predictor to refine subsequent predictions. However, the performance is limited in depth-discontinuous scenes, as illustrated in \cref{fig:fail}. While our method performs exceptionally well in depth-continuous regions (outside the red box), it struggles as the other methods in depth-discontinuous regions (inside the red box, such as trees or rocks). Such depth-discontinuous scenarios present significant challenges for all existing dehazing methods.

\twocolumn[]

{
    \small
    \bibliographystyle{ieeenat_fullname}
   \bibliography{main}
}

\end{document}